\documentclass[times,twocolumn,final,nopreprintline]{elsarticle}

\usepackage[colorlinks=true,breaklinks=true,pdftex]{hyperref}

\usepackage{subfigure}

\usepackage{amssymb}
\usepackage{latexsym}
\usepackage{makecell}
\usepackage{url}

\usepackage{hyperref}
\usepackage{graphicx}
\usepackage{caption}
\usepackage{float}
\usepackage[switch,pagewise]{lineno} 

\newtheorem{definition}{Definition}

\usepackage[T1]{fontenc}
\usepackage[latin9]{inputenc}
\usepackage[table]{xcolor}
\usepackage{collcell}
\usepackage{hhline}
\usepackage{pgf}
\usepackage{xcolor}

\date{}
\begin{document}
\baselineskip 10pt

\begin{frontmatter}

\title{A learning-based approach to feature recognition of Engineering shapes}%
%
\author[1]{Lakshmi Priya Muraleedharan}
\author[1]{Ramanathan Muthuganapathy \corref{cor1}}
\cortext[cor1]{Corresponding author: email: mraman@iitm.ac.in}
\address[1]{Advanced Geometric Computing Lab., Department of Engineering Design, Indian Institute of Technology Madras, India}


\begin{abstract}
In this paper, we propose a machine learning approach to recognise engineering shape features such as holes, slots, etc. in a CAD mesh model. With the advent of digital archiving, newer manufacturing techniques such as 3D printing, scanning of components and reverse engineering, CAD data is proliferated in the form of mesh model representation. As the number of nodes and edges become larger in a mesh model as well as the possibility of presence of noise, direct application of graph-based approaches would not only be expensive but also difficult to be tuned for noisy data. Hence, this calls for newer approaches to be devised for feature recognition for CAD models represented in the form of mesh. Here, we show that a discrete version of Gauss map can be used as a signature for a feature learning.  We show that this approach not only requires fewer memory requirements but also the training time is quite less. As no network architecture is involved, the number of hyperparameters are much lesser and can be tuned in a much faster time. The recognition accuracy is also very similar to that of the one obtained using 3D convolutional neural networks (CNN) but in much lesser running time and storage requirements. A comparison has been done with other non-network based machine learning approaches to show that our approach has the highest accuracy. We also show the recognition results for CAD models having multiple features as well as complex/interacting features obtained from public benchmarks. The ability to handle noisy data has also been demonstrated.
\end{abstract}


\end{frontmatter}


\section{Introduction}

Features in the field of computer-aided design (CAD) or Engineering such as holes, slots, pockets provide semantically higher level information present in a model and have received a lot of attention in the literature.  Significantly, there is no unique definition for a feature but looked at as an entity used in the process of design or manufacturing of a product \cite{SreevalsanS91}.  Pratt defined a feature as: A region of interest on the surface of a part \cite{pratt1988}. Apart from these features, some models may contain interacting features (where one or more features interact with each other). 

 In this paper, the term {\em feature extraction} implies that the portions (or parts) in a model which contribute to a feature are identified. Recognition of a feature implies that the extracted feature is categorised into a through hole, or a blind hole, or a though slot, or a blind slot, or a rectangular pocket etc.

Recognition of features has been an active area of research for almost four decades. One of the first works in this area appear to be from Kyprianou \cite{Kyprianou1980} for the purpose of part-classification followed up by Jared \cite{Jared1984}.  The native format of CAD design data is usually B-Rep and hence most of the works in this field (for e.g., see \cite{Sakurai90} or \cite{Niu201535})) have been motivated in this direction, largely using graph-based approaches. Moreover, 'noise' is usually not present in a B-Rep data.

A mesh is a collection of vertices, edges and faces representing the surface of an object. Though mesh model representation is extensively used in the area of computer graphics, this representation is becoming popular in the area of CAD due to the advent of newer manufacturing techniques such as 3D printing.  Due to digital archiving, many of the data are now acquired though scanning and then then performing a reverse engineering that store  the data in the form of a mesh. The process of scanning can also induce noise in the data. The mesh models also do not intrinsically capture features present in a CAD model.

Quite a few research groups have worked on creating datasets for CAD models. A few popular ones are Engineering Shape Benchmark (ESB) \cite{Jayanti06}, National Design Repository (NDR) \cite{RegliNDS}, GrabCAD \cite{grabcad} etc. ESB has archived around eight hundred plus CAD models in about forty two classes (based on functionality) and NDR has around a few hundred models. However, many of the benchmarks such as ESB store the models in the form of a polygonal mesh, typically as a triangle mesh. 
When the number of vertices and edges become much larger (as is the case in mesh models), direct application of graph-based techniques such as sub-graph matching on the model (that are popular for B-Rep) may not be feasible as it is an NP-Complete problem and not really scalable. Hence,  segmentation of a mesh model, followed by a graph-based approach is employed to recognise them \cite{Xiao2011685,MURALEEDHARAN201851} but the problem of encoding remains.

The encoding of features becomes even more cumbersome if the feature data is large in number and more so when the same has to be verified in a database of hundreds of models.
 
To employ a learning-based approach, a larger number of models (not just in hundreds but at least in thousands) are needed. Recent approaches in the field of CAD also suggest to use learning-based approach (FeatureNet \cite{ZHANG201812}, CADNet \cite{Manda01}, ABC \cite{Koch_2019_CVPR} etc.)

\begin{figure}[!h]
 	\centering
 	\subfigure[\label{fig:cad} ]
 	{\includegraphics[width=3.9cm]{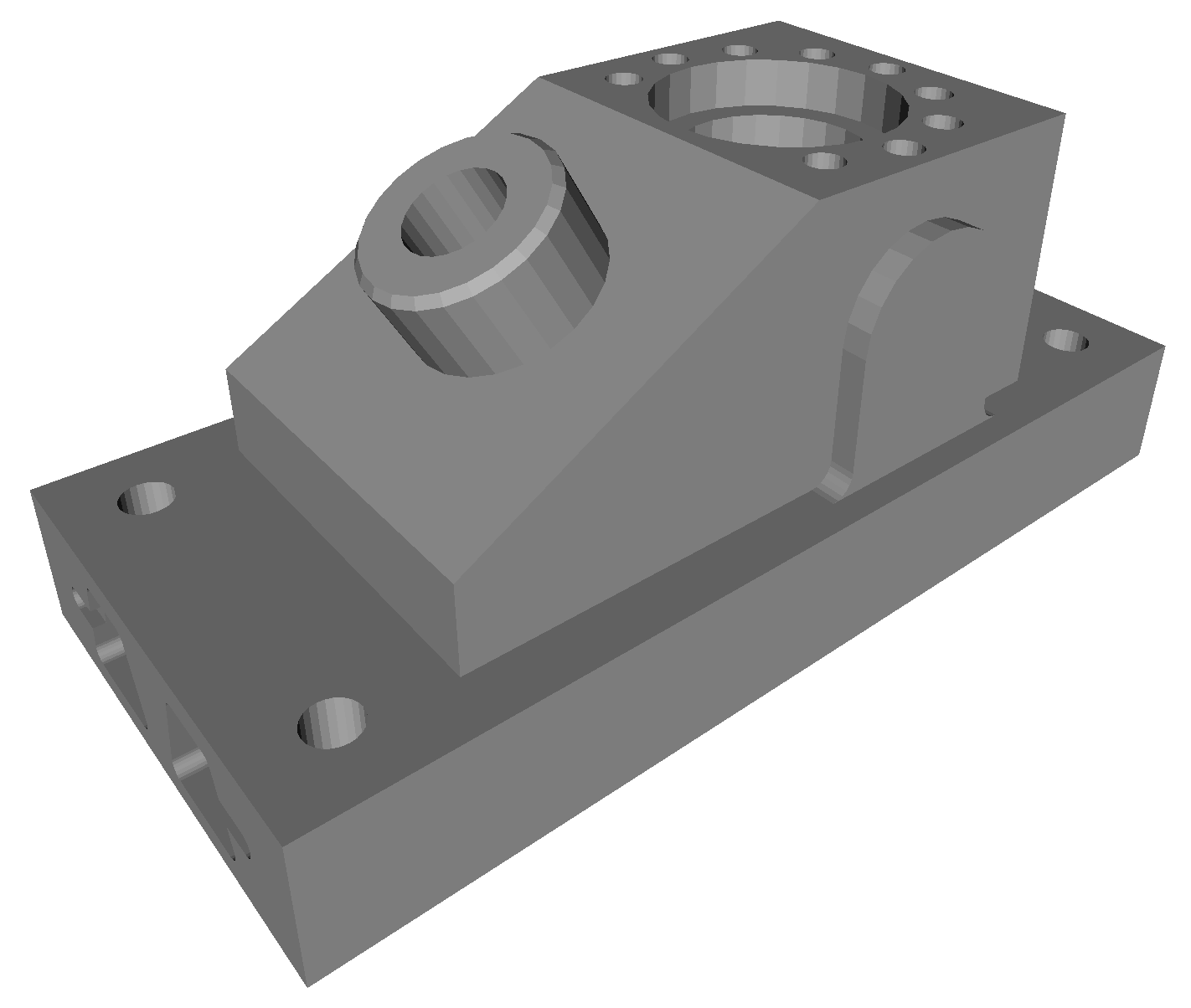}} \qquad
 	\subfigure[\label{fig:ex0201} ]
 	{\includegraphics[width=3.9cm]{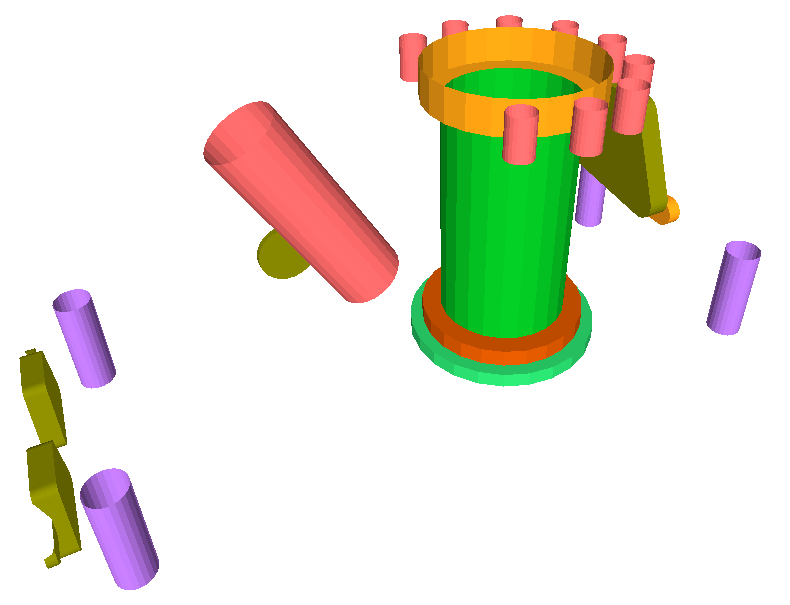}}
 	\caption{ \label{fig:cad1} (a) A CAD mesh model. (b) Extracted features (unrecognised).}
\end{figure}

Zhang et al. \cite{ZHANG201812} proposed deep learning approach using 3D CNN for feature recognition. FeatureNet \cite{ZHANG201812} provides a dataset of individual (single) features having around 24 classes with 1000 models in each of them (augmented with six rotations making the number to 6000 each). Hence, the task of training assumes that a model is made of a single feature. In practise, a CAD model will comprise of multiple features (as in Figure \ref{fig:cad}) and hence to recognise them, features have to be extracted first (such as the ones in Figure \ref{fig:ex0201}). In FeatureNet \cite{ZHANG201812}, to extract features in a model with multiple features as in Figure \ref{fig:cad}, they resort to voxel-based labelling \cite{Vanderwalt2014} along with watershed segmentation \cite{6976891}. 

In general, voxel-based 3D CNN is usually memory intensive and results in a long time for training the data. It also requires high number of hyperparameters to be tuned and hence much more time consuming than ML techniques. The training time can be significantly reduced if we chose an appropriate signature to encode a feature. It is also very important that a signature can be represented concisely, requiring only very little memory. The signature should also be rotation and translation invariant to capture the feature information. 

In this paper,  a  machine learning approach to recognise a feature by encoding through discrete Gauss map representation similar to the one described in \cite{EGI}. We show that our approach can achieve similar accuracy as that of 3D CNN but with much less training time and memory consumption. Number of hyperparameters to be tuned is also lesser than the ones that use network-based approaches such as ANN/CNN. After training each feature, our recognition approach is then applied on models with multi-features from datasets such as ESB and NDR and the results are discussed. The following are the key contributions of the work;
\begin{itemize}
\item Discrete Gauss map based encoding of each feature, thereby facilitating a concise representation.
\item Propose to use classical machine learning approach that uses less memory and runs much faster.
\item Complex and interacting features can be handled using this approach. 
\item We show that our approach can also handle noise in the data.
\end{itemize}

The rest of the paper is organized as follows: Section \ref{sec_rw} presents the related works to feature recognition under various inputs.  Section \ref{sec_method} discuss the methodology with details of feature extraction (section \ref{sec_featureex}, discrete Gauss map as shape signature (Section \ref{sec_sig}), choosing a machine learning model (Section \ref{sec_mlmodel}) and feature recognition (Section \ref{sec_fr}). Results are presented for single features, multi-feature models, complex/interacting features, unseen feature and noisy inputs and discussed in  Section \ref{sec_resdis}. Section \ref{sec_conc} concludes the paper.  

\section{Related work} \label{sec_rw}

\subsection{B-Rep input}
Initial work in the area of feature recognition started for boundary representation (B-Rep) models. Graph-based method is a very prominent approach for B-Rep models using sub-graph matching \cite{Joshi:1988}, an  NP-complete problem. Approaches based on artificial neural networks (ANN) have also been employed in several works \cite{PRABHAKAR1992381,NEZIS1997523,LANKALAPALLI1997,Onwubolu1999,Sunil2009ANN}.

\subsection{Mesh model as input}

When a mesh model is taken as input, a few of the approaches use clustering to extract basis primitive shapes such as plane, sphere or cylinder \cite{Attene:2006:HMS, ThinPlate2010, mortara2004blowing, mortara2004plumber}. Though these approaches can extract certain shapes (features), they cannot be used for recognition of different kinds of mechanical features. 

In another approach, termed as Geometry-based ones, they try to identify the feature lines in the body and then subsequently use graph-based connectivity to extract the features \cite{owen2001mesh, jiao2002feature,vidal2011robust, zhang2001efficient,chen2006efficient,wang2012effective,Sunil2008}. Though slicing using planes has been employed in \cite{Nepal2016}, the contours are analysed using a threshold on the angles. A non-parametric approach based on random cutting planes \cite{MURALEEDHARAN201851} can extract a lot of features but the geometry-based approach to recognition limits the types of features. 

\subsection{CNN-based approaches}

Within the last few years, a large body of literature is available in the areas of both images and graphical models and citing all of them is beyond the scope of this paper. Nevertheless, in the traditional CAD/Engineering field, there are only a few works employing machine learning (ML). Balu et al. \cite{BaluLYKS16} have developed a voxel-based 3D CNN approach to determine if a design is suitable for manufacturing. For recognising drilling features, a voxel-based 3D CNN approach for B-Rep models has been proposed in \cite{GHADAI2018263}).   A big CAD model dataset \cite{Koch_2019_CVPR} has about millions of models but no classification details appear to be available. CADNet \cite{Manda01} has proposed a dataset of CAD models and used 2D CNN for their classification. Recently, Zhang et al. \cite{ZHANG201812} proposed FeatureNet, a voxel-based 3D CNN approach to learn machining or manufacturing features and recognise them. However, a multiple feature model needs to be segmented and then each segment is recognised into a particular feature type using the proposed 3D CNN. View-based deep-learning framework has been proposed in \cite{Shi2020}, where features are segmented first and then recognised. In the multi-view approach \cite{Shi2020}, the number of views also need to be experimentally identified in addition to the hyperparameters  of the deep network. 

It is evident from the various prior works, very less amount of work has been done in the field of CAD models using a learning-based approach. Further, in the research of feature recognition of CAD models, the works used deep networks, which is both time and memory consuming. In this paper, it is shown that, rather than using deep networks, the problem can be addressed using a non-network ML-based approach. The details of the approach, as well as the advantages/disadvantages with respect to 3D CNN-based approach, are also discussed in detail in the subsequent sections of this paper. 

\section{Methodology} \label{sec_method}

The overall approach consists of the following steps: \\
1) Feature extraction and alignment. \\
2) Shape signature using discrete Gauss map. \\
3) Choosing a machine learning model. \\
4) Feature recognition using training and testing of the ML model \\

\subsection{Feature Extraction} \label{sec_featureex}
The features from a mesh model are extracted first. In this paper, rather than the modeling of individual features, we employ a feature extraction method to extract each volumetric feature from a mesh model. This is possible because the recently built dataset called FeatureNet \cite{ZHANG201812} has at least one-model-per-one-feature. We use one model from each class in FeatureNet to extract each feature. Though there are several approaches to extract a feature, such as segmentation-based ones \cite{zhang2001efficient}, many of them require user input or threshold. Hence we may have to manually pre-process those surfaces into features or not which is time consuming. The approach presented in \cite{Nepal2016} uses cutting planes but still uses angle threshold as a user parameter. The random cutting plane approach presented in \cite{MURALEEDHARAN201851} is completely automatic and did not require any user intervention and hence we use this approach. 

\textcolor{black}{One disadvantage of the method presented in \cite{MURALEEDHARAN201851} is its inability to extract open slot structures since the cutting planes cannot extract closed inner loops in such structures. Hence we have devised a mechanism to automatically extract such open slot structures using the concept of hyperbolic points.}

\begin{figure}[!h]
\begin{minipage}{0.48\linewidth}
	\centering
	\includegraphics[width=\linewidth]{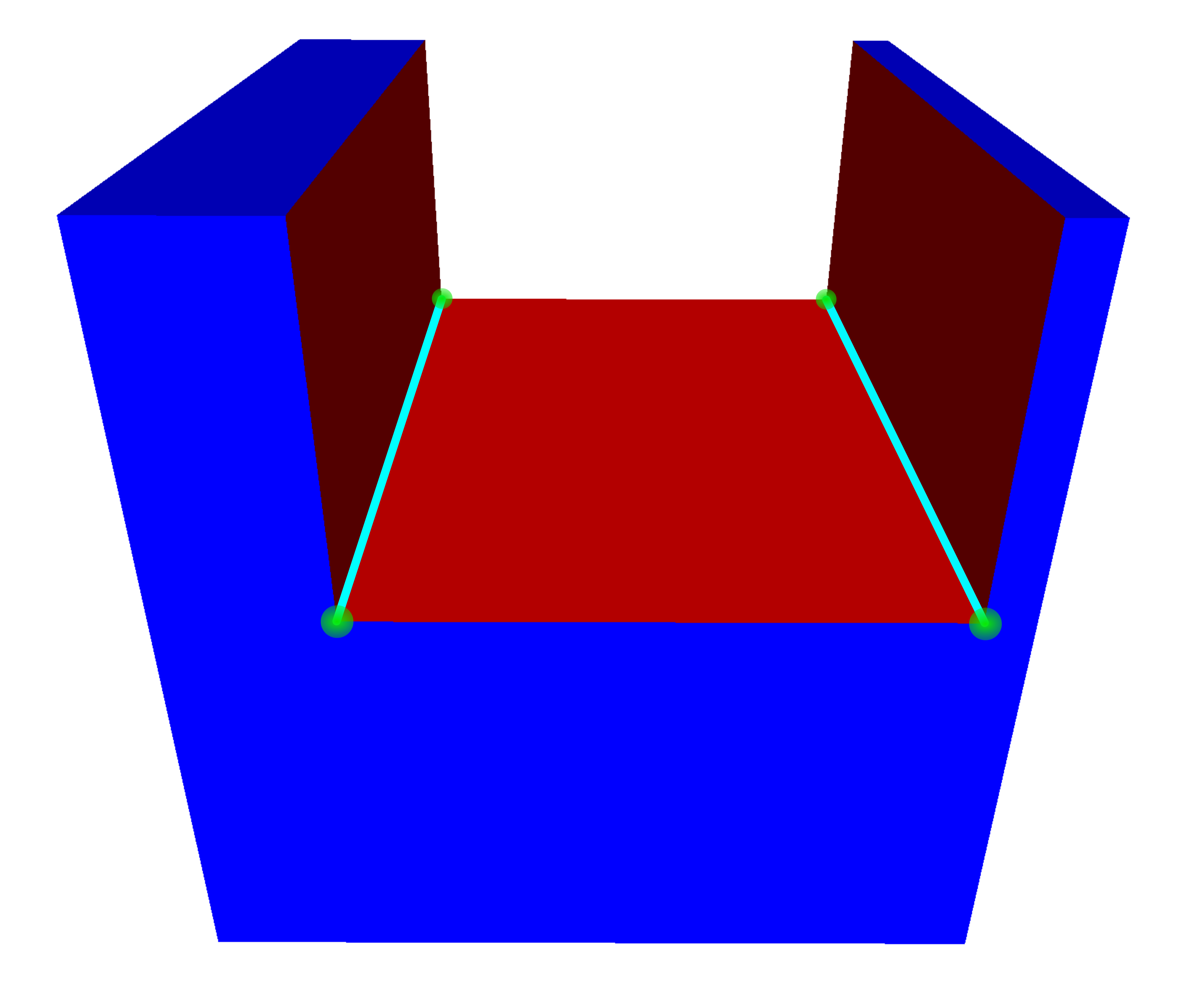}
	\caption{\label{fig:openslot} Open slot (red). Hperbolic vertices (green) and edges (cyan).  }
\end{minipage}
\begin{minipage}{0.5\linewidth}
	\centering
	{\includegraphics[width=0.9\linewidth]{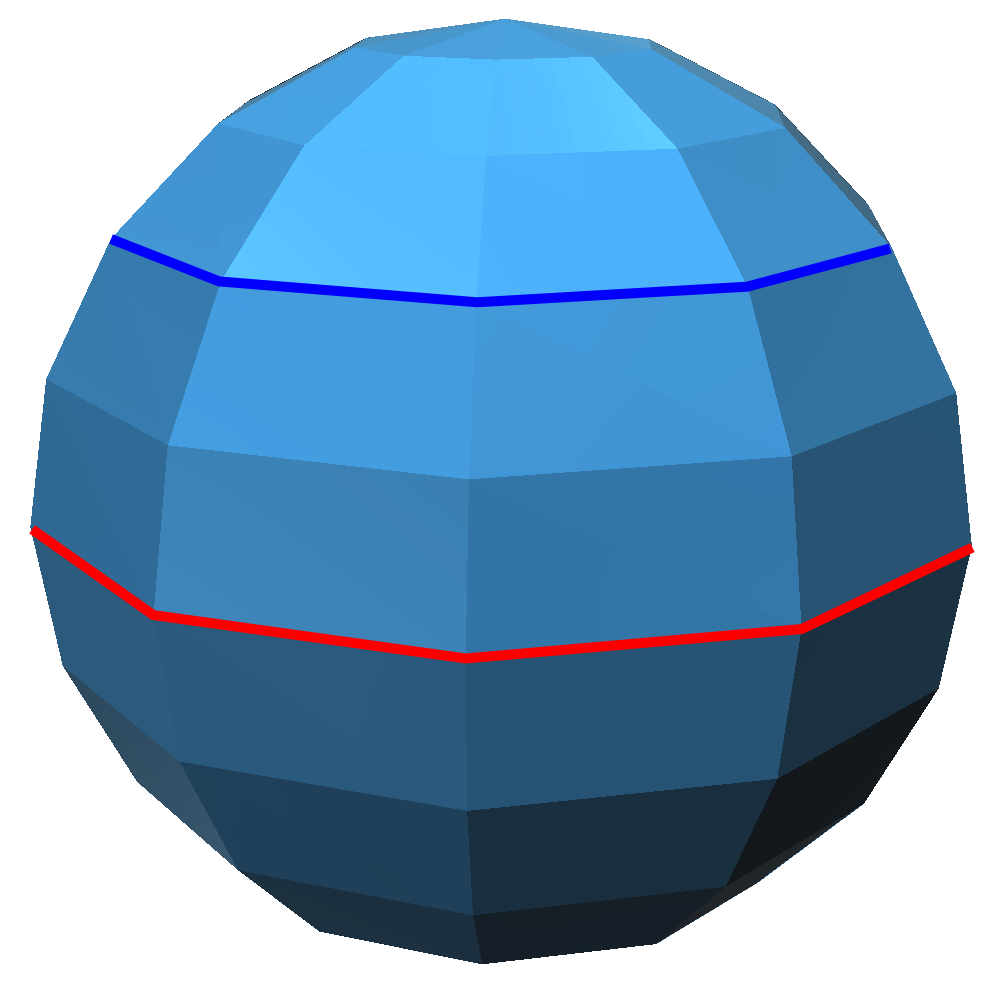}}
	\caption{\label{fig:discGauss} Discrete Gauss map.}
\end{minipage}
\end{figure}

\begin{definition}
Let $dN_p:T_{p}(S)\rightarrow T_{p}(S)$ be the differential of the Gauss map, where $T_{p}$ is the tangent plane. The determinant of $dN_{p}$ is the Gaussian curvature $K$ of $S$ at $p$ \cite{DBLP:books/daglib/0090942}. In terms of principal curvatures,
\begin{equation}
K = k_{max}.k_{min}
\end{equation}
\end{definition}

\begin{definition}
\label{hyperbolic}
A point p of a surface S is called Hyperbolic if $det(dN_p) < 0$ \cite{DBLP:books/daglib/0090942}
\end{definition}
 
\begin{enumerate}
    \item Concave edge ($E_C$): An edge is concave if the angle between two of its incident triangles is less than $\pi$ (angle measured from the outside). 
 	\item Hyperbolic vertex ($V_H$): A vertex $V$ with $K<0$ in the input mesh model (as shown in Figure \ref{fig:openslot} with green dots).
 	\item Hyperbolic edge ($E_H$): A concave edge connecting two hyperbolic vertices in the mesh model (as shown in Figure \ref{fig:openslot} in cyan).
 \end{enumerate}

\textcolor{black}{At a hyperbolic point, the Gaussian curvature is negative, since the principal curvatures have opposite signs. From Figure \ref{fig:openslot}, we can view that the red facets are the facets which makeup the rectangular open slot. The green vertices are hyperbolic vertices and the concave edges which connect the hyperbolic vertices are hyperbolic edges. It can be noticed that whenever an open slot structure is carved on a surface, there will be a presence of hyperbolic edge. Hence we can use the hyperbolic edges to identify such regions.}

\textcolor{black}{We consider the hyperbolic edges on the surface and take its both incident facets $f_1$ and $f_2$ and use them to construct supporting planes $P_1$ and $P_2$. Let $F$ be a set of facets of the mesh lying on such supporting planes. If a connected component of the set $F$ is extracted then each component $c_i$ will be an independent feature. Thus, open slots can be extracted by this method. }

\subsubsection{Alignment} \label{sec_align}

An extracted feature could be aligned in various orientations in its original model. A uniform representation of each feature has to be found to make them orientation invariant. In order to transform each feature, we calculate the oriented bounding box of each feature and find the longest axis of this box. \textcolor{black}{The oriented bounding box is found from the covariance matrix of the convex hull facets of the feature. Compute the Eigen vectors of that matrix and find the extent of the model along these directions to fit the bounding box. Using the convex hull to find the major axis will avoid the effect of small variations in the inner vertices from affecting the alignment}. The angle $\theta$ between the longest axis of the bounding box and Z axis is found and the object is rotated by $\theta$ thereby aligning the major axis of each part to Z axis. 
The aligned model might face away from or towards the origin after this processing. In order to preserve uniformity, the feature boundary is always made to face towards the origin. If there are two boundaries, then the largest outer boundary is made to face the origin. 

\subsection{Shape signature using discrete Gauss map} \label{sec_sig}

In order to recognise a feature, there can be two components, viz., geometry and topology. For example, a feature can be cylindrical in geometry but topologically, it can be classified into either a through hole or a blind hole. In order to recognise geometry such as cylindrical, conical etc. which typically fit into primitives, Gauss map has been proven to be a good representation \cite{Xiao2011685}.

\begin{definition}
Given a surface $X$ lying in $R^3$, the Gauss map is a continuous map $N:X \rightarrow S^2$ (unit sphere) such that $N(p)$ is a unit vector orthogonal to $X$ at $p$, namely the normal vector to $X$ at $p$. 
\end{definition}

Geometrical shapes such as cone and  cylinder is mapped to circles in a Gauss map. However, such a circle-based approach on a Gauss sphere can handle only a limited number of geometries. For detecting topology of a feature, connectivity information may be used, such as connected components from a graph. However, encoding all different ways of connected components will be cumbersome. 

\subsubsection{Discrete Gauss Map}
\label{sec_Gauss}

Rather than using circle-based Gauss map approach, the key idea is to use a discretized Gauss map as follows: \textcolor{black}{In the discrete version of the Gauss map, Gauss sphere is sampled with a certain number of vertices (NV) as shown in Figure \ref{fig:discGauss}. Every facet normal in the extracted feature is then mapped to a vertex normal in discretised Gauss map using a k-d tree approach and nearest neighbor algorithm.} A $[NV\times1]$ Gauss signature is created for every feature which denotes the Gauss sphere vertices latitude wise. When a facet normal is mapped to a vertex normal, the area of the facet is added to its corresponding position in the vector. Finally, the Gauss signature is normalized to get a representation which gives the percentage of the area on the surface oriented to each direction \textcolor{black}{which will capture the topology of the feature}. Every feature is thus represented with a $[NV\times1]$ vector. 

\begin{figure*}[h]
	\centering
	\includegraphics[width=\linewidth]{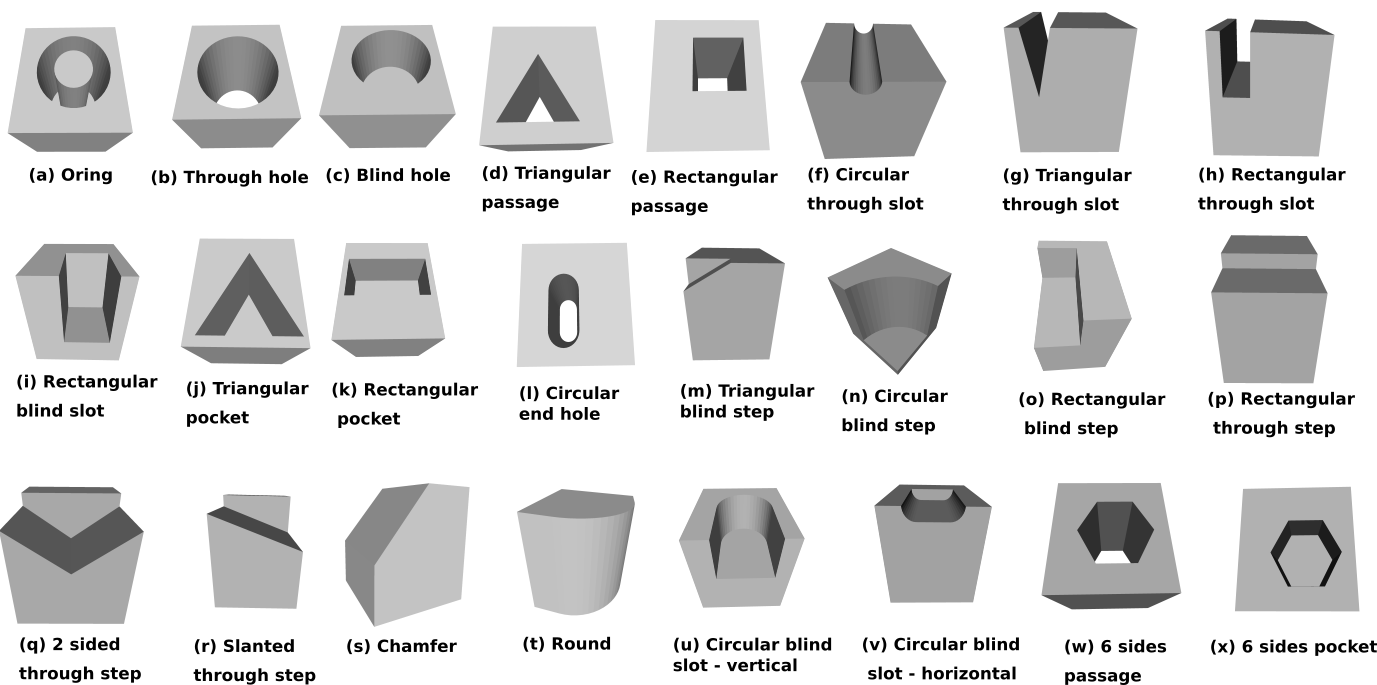}
	\caption{\label{fig:featureclass} Various categories of Features \cite{ZHANG201812} }
\end{figure*}

\textcolor{black}{In general, this signature is similar to the Extended Gaussian Image (EGI) \cite{EGI}, which uses a face-to-face mapping whereas we propose face-to-vertex mapping. We also use as `weights' when multiple normals map onto the same vertex which the EGI does not take into account.} Figure \ref{fig:featureclass} shows a set of models having features that are either different in geometry or topology or both along with each of the labels \cite{ZHANG201812}. For example, the circular end hole (Figure \ref{fig:featureclass}(l)) has different geometry and topology than that of the blind hole (Figure \ref{fig:featureclass}(c)). The proposed discrete Gauss map enables to handle such wide variety of features (note the representation of each feature is in the form of a mesh and not B-Rep).

\subsubsection{Choosing a machine learning model} \label{sec_mlmodel}

There are several classifiers available such as support vector machines (SVMs), Decision Trees, Random Forests etc. As our feature recognition problem is inherently a multiclass classification problem, either decision trees or random forests is more suitable. As random forests build multiple decision trees and merge them together to get a probability value of belonging to a class, we use random forests as the classifier (also see Section \ref{sec_compclassifiers} for comparison). 

\subsection{Random Forest} \label{sec_randforest}

Random Forest \cite{Breiman2001} is a classical machine learning algorithm proposed by L Breiman. Random forests are tree predictors combined in such a way that each tree is a randomly sampled shape signature from the same distribution for each tree in the forest. Random forest is also robust with respect to noise.

\begin{definition}
A random forest is a classifier consisting of a collection of tree-structured
classifiers ${h(x, \Theta_k ), k = 1,...}$ where the ${\Theta_k }$ are independent identically distributed random vectors and each tree casts a unit vote for the most popular class at input x. \cite{Breiman2001}
\end{definition}

Not every tree sees all the features or all the observations, and this guarantees that the trees are de-correlated and therefore less prone to over-fitting. Each tree is also a sequence of yes or no questions based on a single or combination of shape signatures.

The input vector $[NV\times1]$ is fed into the random forest and the output is considered for the final labeling.

\begin{figure*}[!t]
	\centering
	\includegraphics[width=\linewidth]{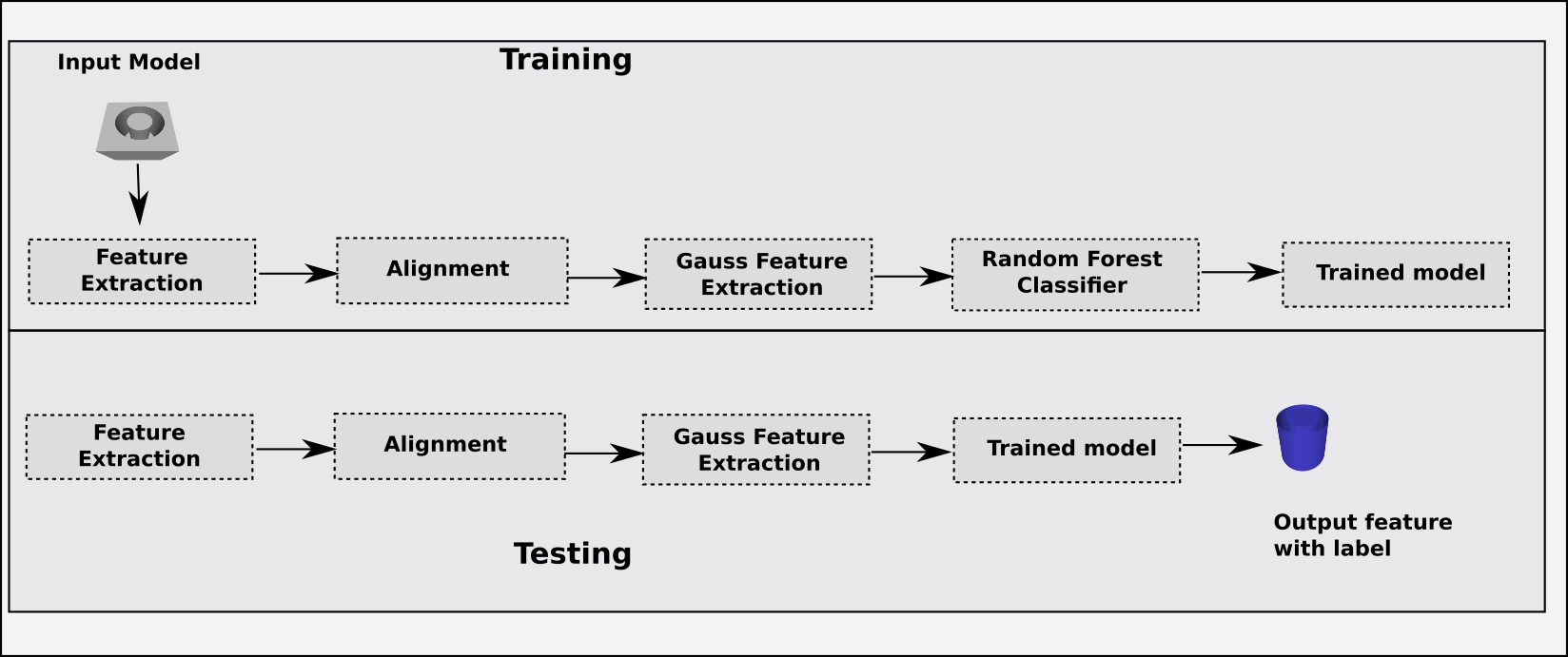}
	\caption{\label{fig:blockdiag} Block Diagram of the overall methodology of feature using the proposed machine learning approach.}
\end{figure*}

\subsection{Feature recognition using training and testing of the ML model } \label{sec_fr}

The overall feature recognition using training and testing of the ML model  is indicated in Figure \ref{fig:blockdiag} and divided into training and testing phase. The approach starts with feature extraction (Section \ref{sec_featureex}), followed by alignment (Section \ref{sec_align}) and then the vector $[NV\times1]$ was generated using the Gauss map discretization (Section \ref{sec_Gauss}). The vector was then trained using the classifier as discussed in Section \ref{sec_randforest}.

\begin{figure}
\begin{minipage}{0.48\linewidth}
	\centering
	\includegraphics[width=\linewidth]{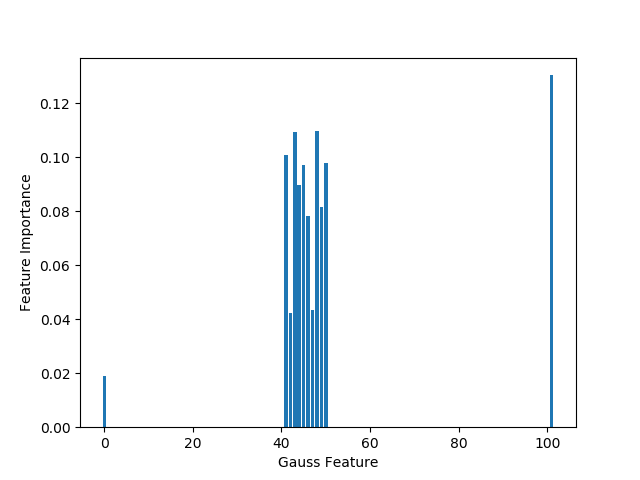}
	\caption{\label{fig:featureimp}  Signature Importance of Gauss feature. }
\end{minipage}
\begin{minipage}{0.5\linewidth}
\centering
	\includegraphics[width=\linewidth]{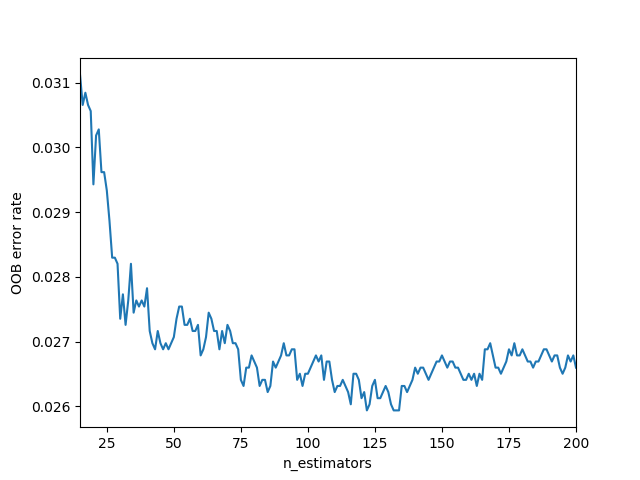}
	\caption{\label{fig:error} Number of estimators vs error rate}
\end{minipage}

\end{figure}

\subsubsection{Training, validation and testing}

In order to perform a single feature recognition, we use the 24 classes (Figure \ref{fig:featureclass}) from the dataset \cite{ZHANG201812} which consists of blocks with one feature in each block. Each category consists of 1000 examples in each. Hence, we used only 24,000 as opposed to 144,000 models as we did not use their data augmented models. The training, validation and test set were split in a ratio of 70:15:15. Later the feature extraction is performed to extract the feature of each model. The vector from the Gauss map discretization is then obtained from each of the features. The training set is then used to train the Random forest model. 

\subsubsection{Gauss map discretization}

After several experiments, the vector size for the Gauss map discretization is fixed as $[102\times1]$. This was done after looking at the signature importance histogram (Figure \ref{fig:featureimp}). We can visualize the most important indices of the extracted signature. It is very clear that the most important one is the final 102 signature which lies on the pole of the Gauss sphere, that will distinguish between a through hole and a pocket. The other important values are lying on the equatorial plane of the Gauss sphere since the alignment of most of the faces in a feature is oriented in that direction. This plot suggests which indices of the signature are important and which indices are not. Looking at the signature importance can give a direction of which of the variables have the most effect in the model. 

For example, consider a rectangular passage (Figure \ref{fig:featureclass}(e)) and a rectangular blind slot (Figure \ref{fig:featureclass}(i)). The most differentiating factor between both models is the presence of the bottom face in a rectangular blind slot. Since both are oriented, the normal of the closed face in the blind slot will be facing the south pole of the Gauss map. In such cases, the differentiating factor will be the final 102 index. Hence training on all such features will increase the weight of the final index (which signifies the south pole) to the highest value. The other higher values occur in the indices 40 to 50 which indicates that they lie in the equatorial plane of the Gauss map. Since the features are aligned, most of the facet normals will lie on the equatorial plane and hence the reason for higher value for signature importance.

\begin{figure*}[!t]
	\centering
	\includegraphics[width=\linewidth]{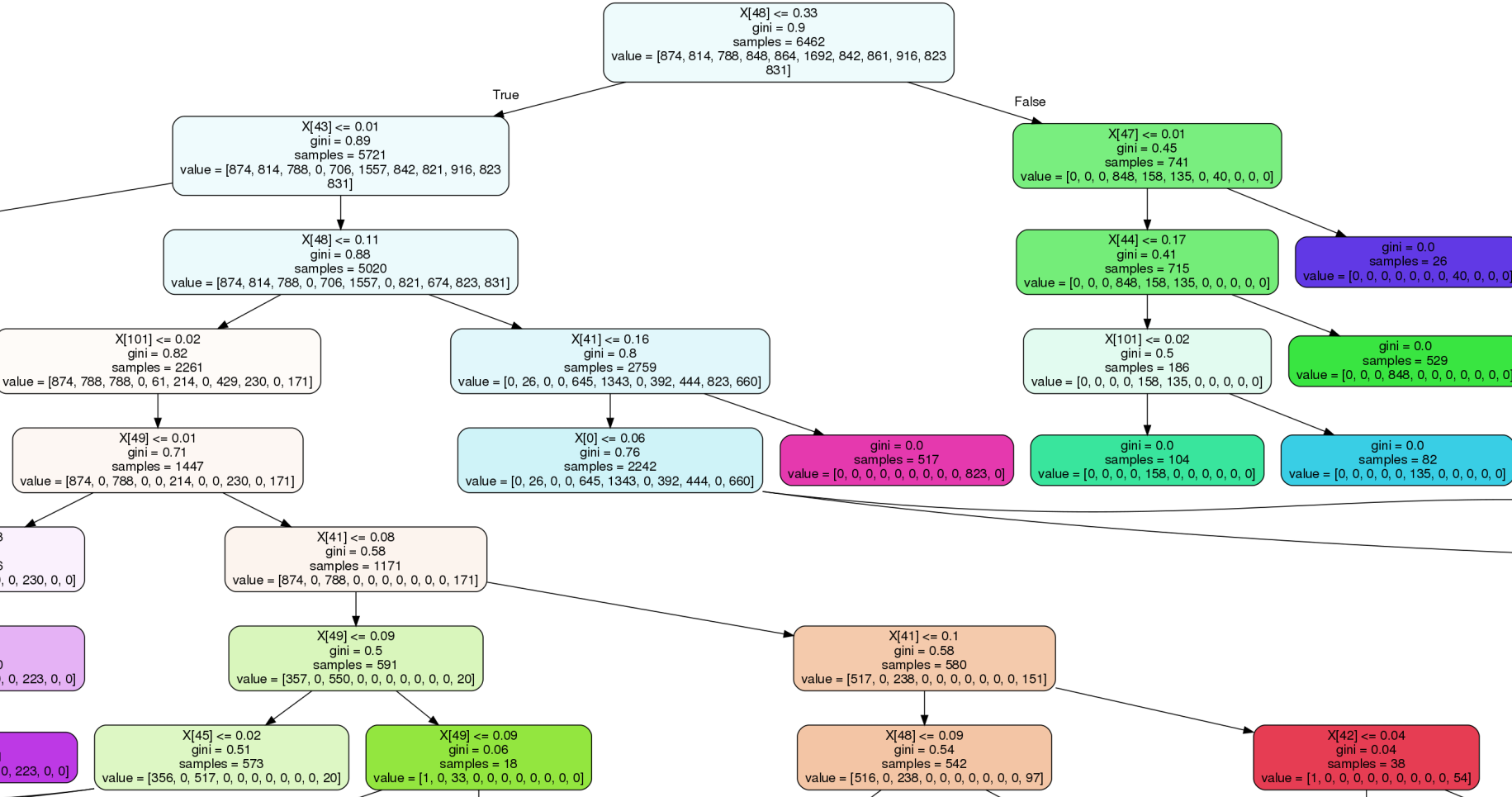}
	\caption{\label{fig:randomforest} A section of one of the tree from our trained Random forest}
\end{figure*}

\subsubsection{Hyper-parameter setting in random forest and system configuration}

The random forest model hyper-parameters such as estimators and max depth were finalized by experimenting with the number of estimators as shown in Figure \ref{fig:error}. Estimators are the number of trees the forest will contain. Max depth is the maximum depth the tree can attain. The number of estimators vs the error was plotted and the value which gave the least error was chosen. Similarly depth parameter was chosen by experimentation. The parameters used for the random forest model is number of estimators=130, max depth=100.

\begin{figure*}[!h]
	\centering
	\includegraphics[width=\linewidth]{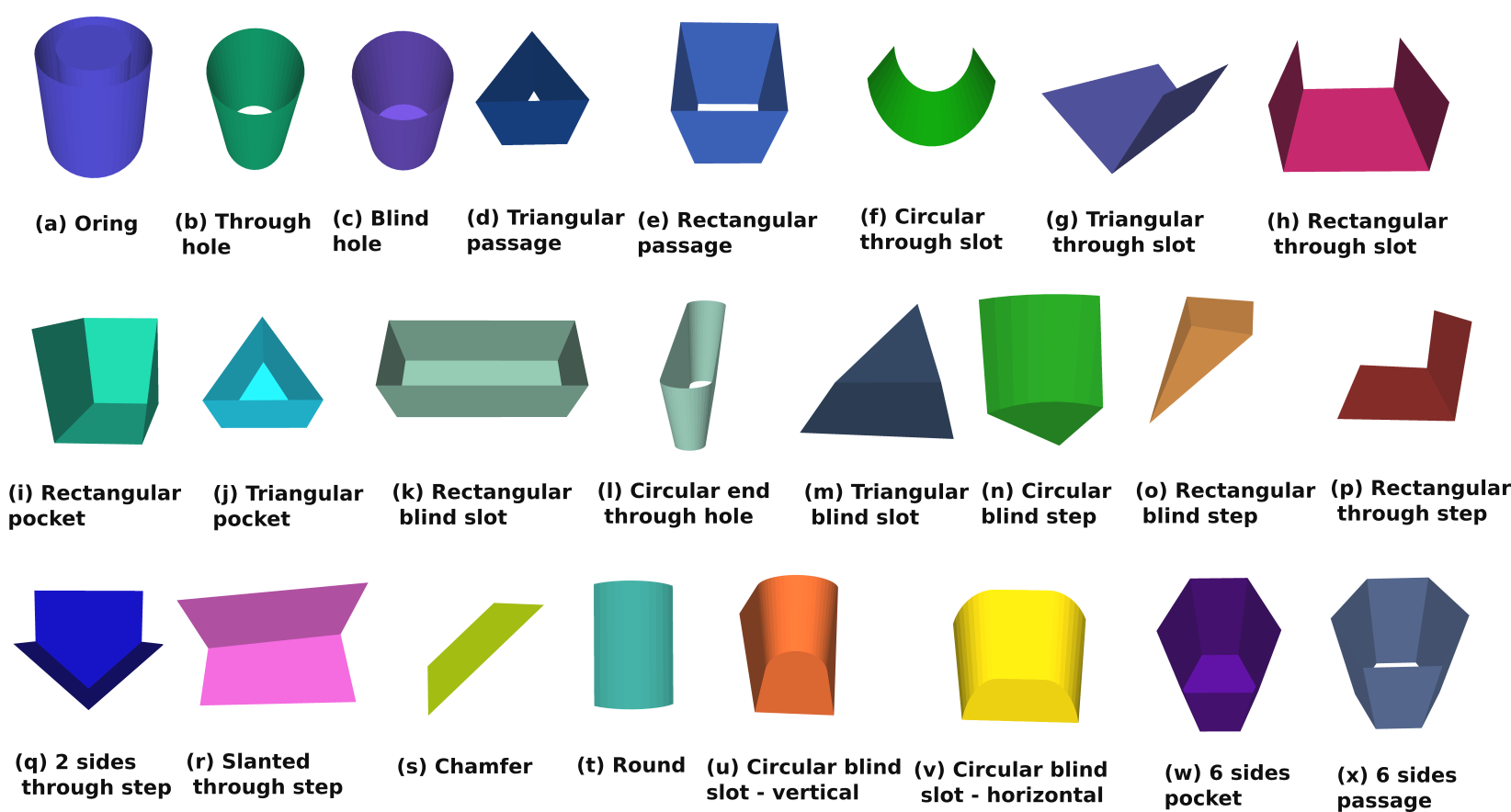}
	\caption{\label{fig:featuresingle} Single feature recognition.}
\end{figure*}

The feature recognition using random forest has been implemented using Python 3 and Scikit-learn \cite{scikit-learn}. \textcolor{black}{Figure \ref{fig:randomforest} shows a section of one of the trees from our trained classifier. Each node in the random forest tree denotes the decision made with respect to the feature and other details such as the number of samples under each split, gini impurity etc. Only a section of one of the trees has been displayed due to the space constraints.}

 All the implementations were carried out on a system wth Ubuntu 18.04 Operating System. The system has an Intel Xeon CPU with 32GB RAM.
 
\section{Results and Discussion} \label{sec_resdis}

 \subsection{Single feature recognition}

The results of the recognition for the features in Figure \ref{fig:featureclass} are shown in Figure \ref{fig:featuresingle}. All the features have different colors indicating that each of them has been recognised into its own class (the names indicate the class). 

\subsubsection{Performance}

It may be noted that the dataset of models is split into 70:15:15 for training, validation and testing. The training accuracy was 100\%, the maximum possible value. During testing, the accuracy dropped a little bit to 97.90\%. Running Time for training and testing the features was 3.19s.

\begin{figure}[!h]
	\centering
	\subfigure[\label{fig:ring _misclass} Input model of Ring feature]
	{\includegraphics[width=.2\linewidth]{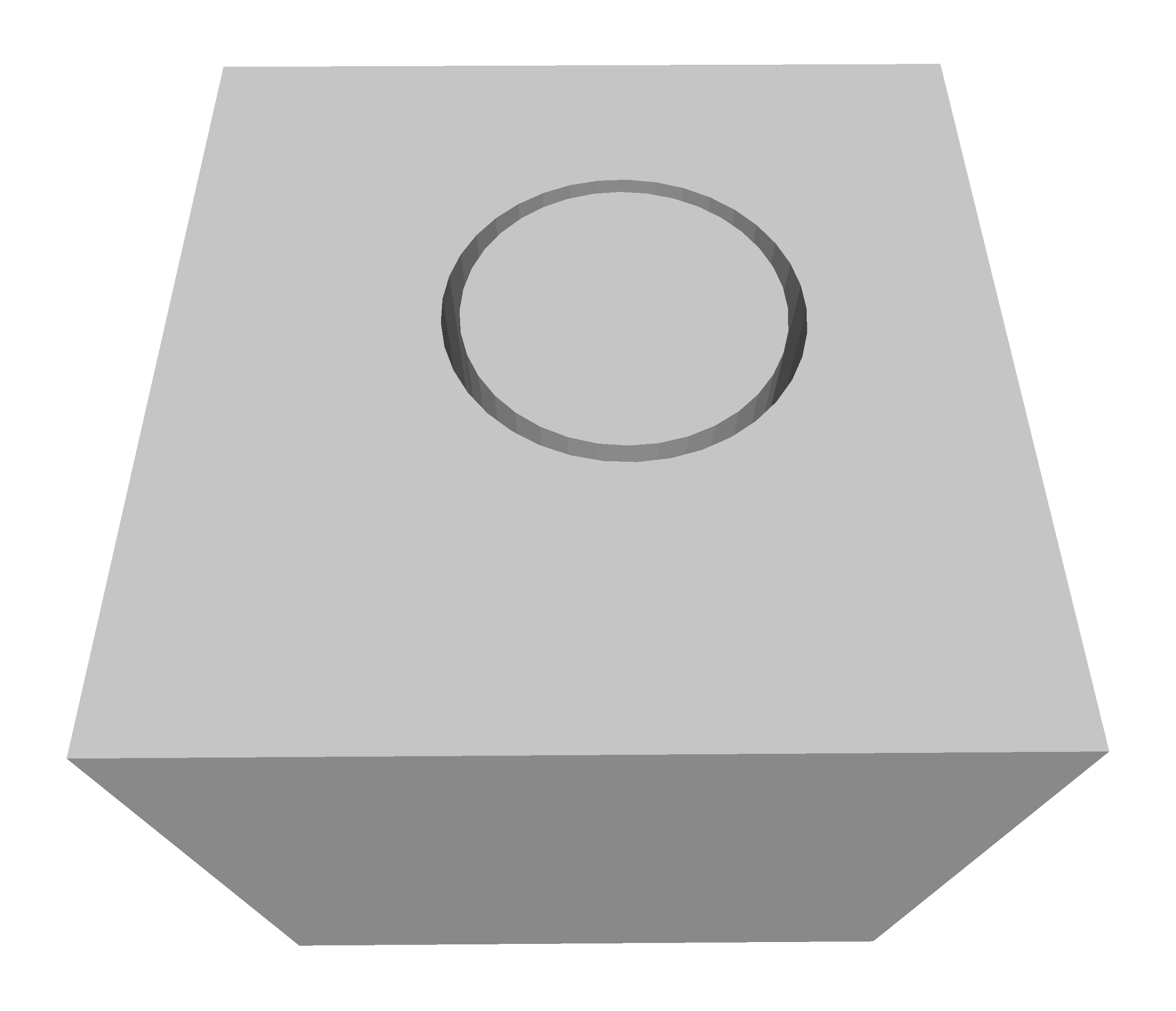}}\qquad
	\subfigure[\label{fig:0_57300} Ring feature extracted (top view)]
	{\includegraphics[width=.2\linewidth]{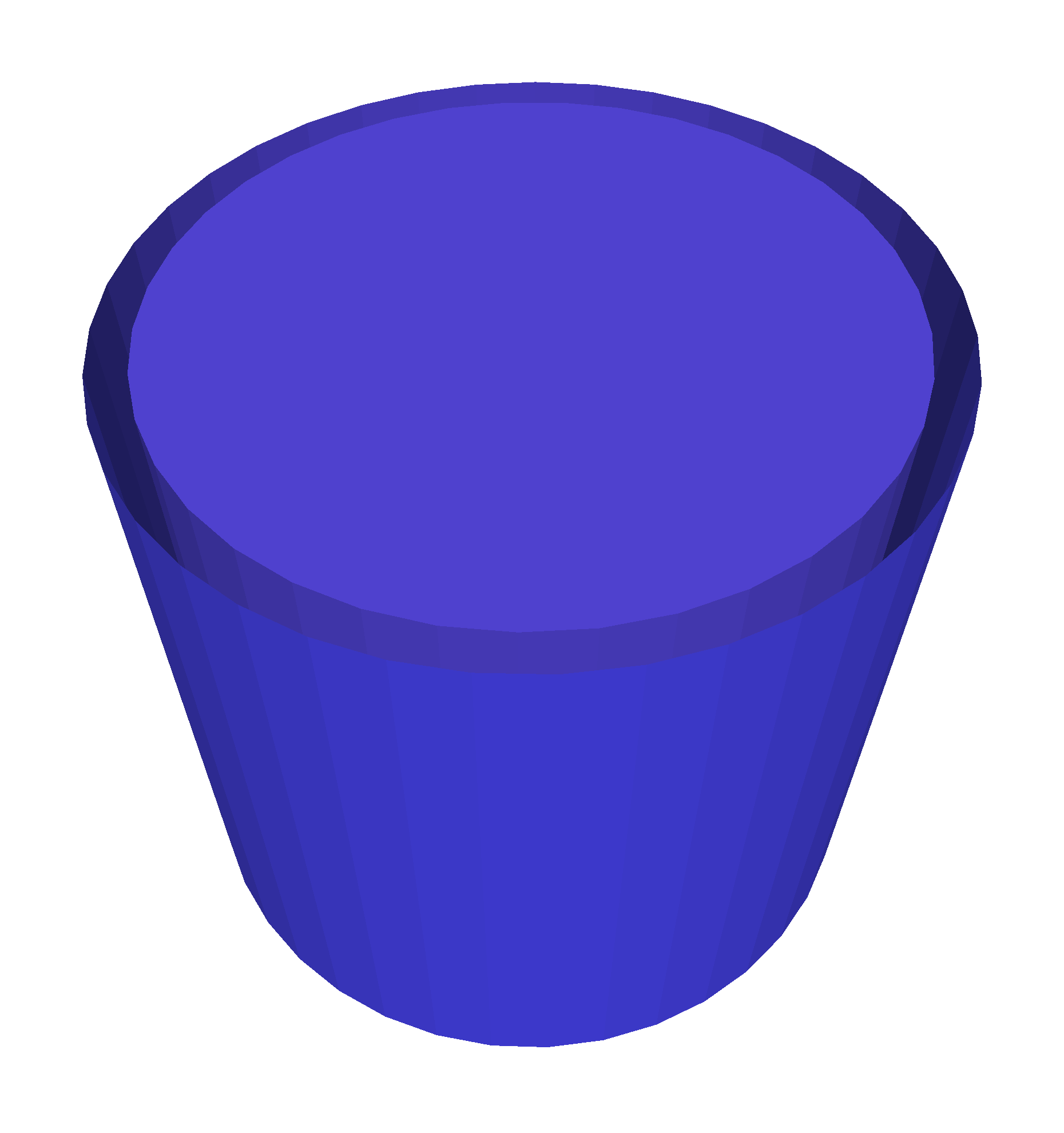}}\qquad
	\subfigure[\label{fig:0_57301} Ring feature extracted (bottom view)]
	{\includegraphics[width=.2\linewidth]{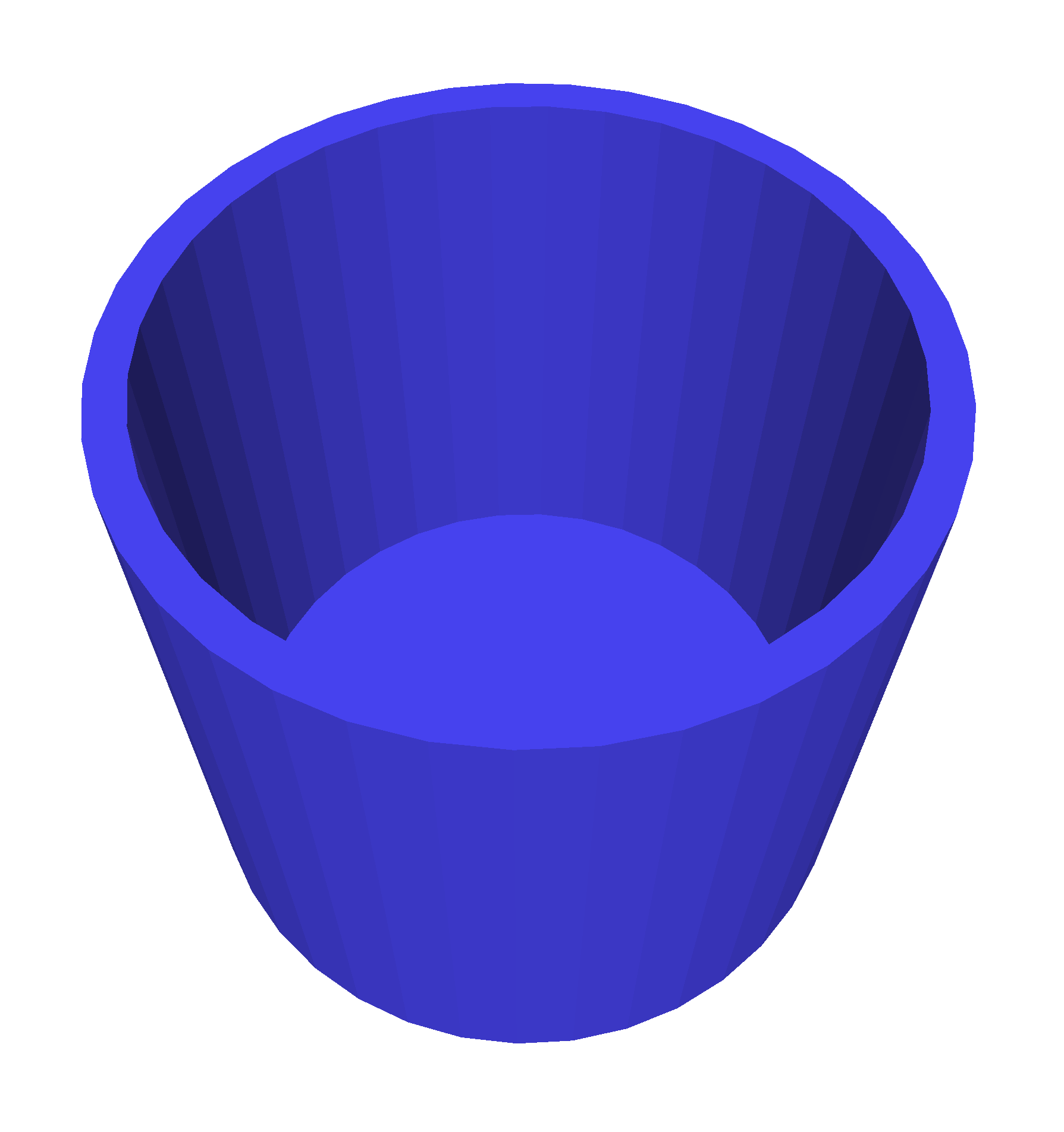}}
	\caption{\label{fig:ring_misclass} Case of model with ring feature similar to blind slot causing mis-recognition}
\end{figure}

We have noticed that the highest rate of misrecognition occurs between a ring and a blind hole. The reason for this misrecognition is due to the structural similarity between the ring feature and blind hole feature when extracted. In some of the models for testing, which were misrecognised, we have noticed that when the width of the ring is very less, the model structure is quite similar to that of a blind slot. Figure \ref{fig:ring_misclass} denotes one of such ring structures which causes this misrecognition. \textcolor{black}{ Figures \ref{fig:0_57300} and \ref{fig:0_57301} are the top and bottom view of the feature which shows the similarity with the blind slot.}

 \begin{figure*}[!h]
    \begin{minipage}{\linewidth}
        \centering
        \subfigure[\label{fig:268_in} Anchor input.]
        {\includegraphics[width=.2\linewidth]{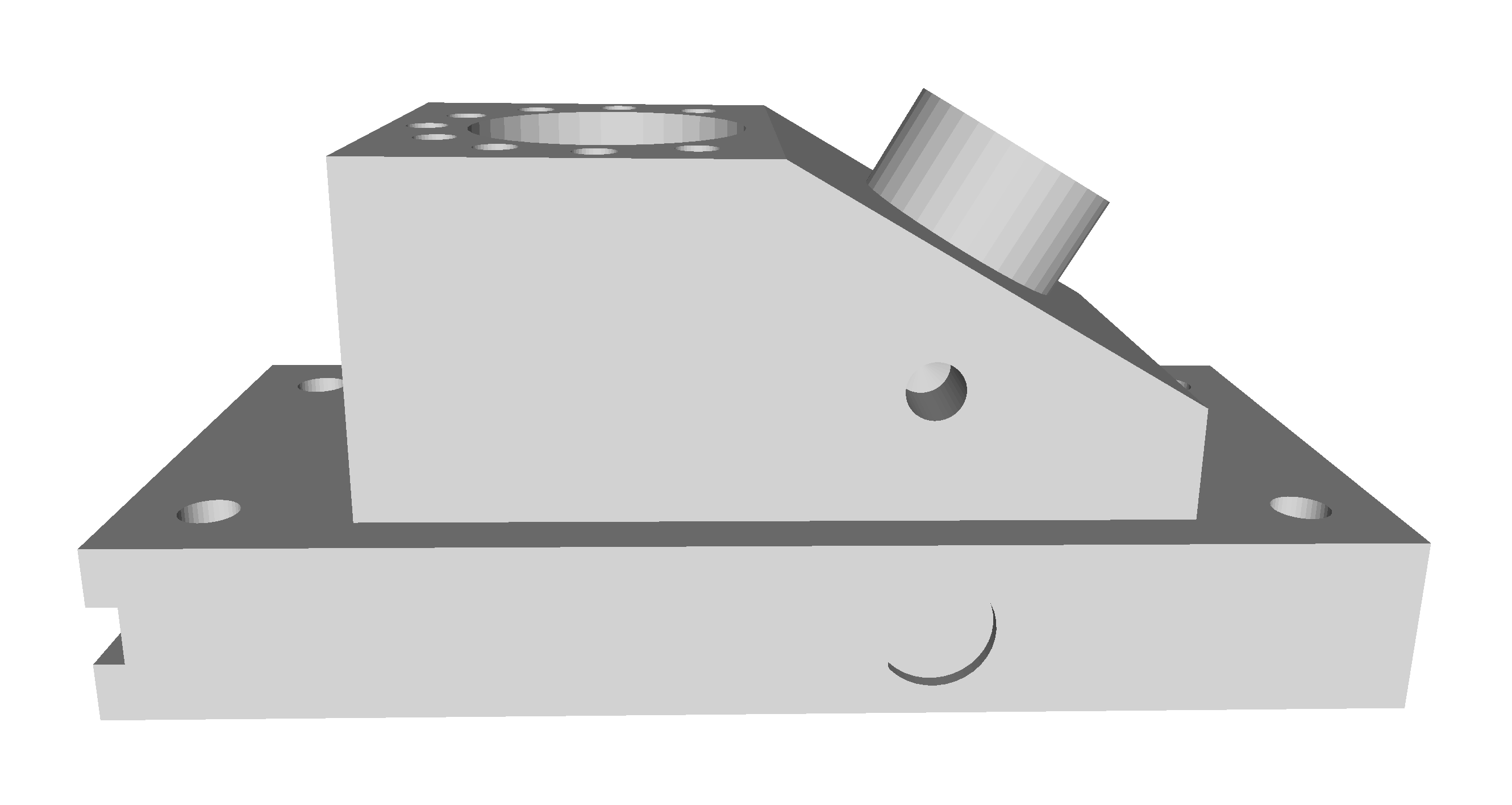}} 
        \subfigure[\label{fig:268_out} Anchor features.]
        {\includegraphics[width=.2\linewidth]{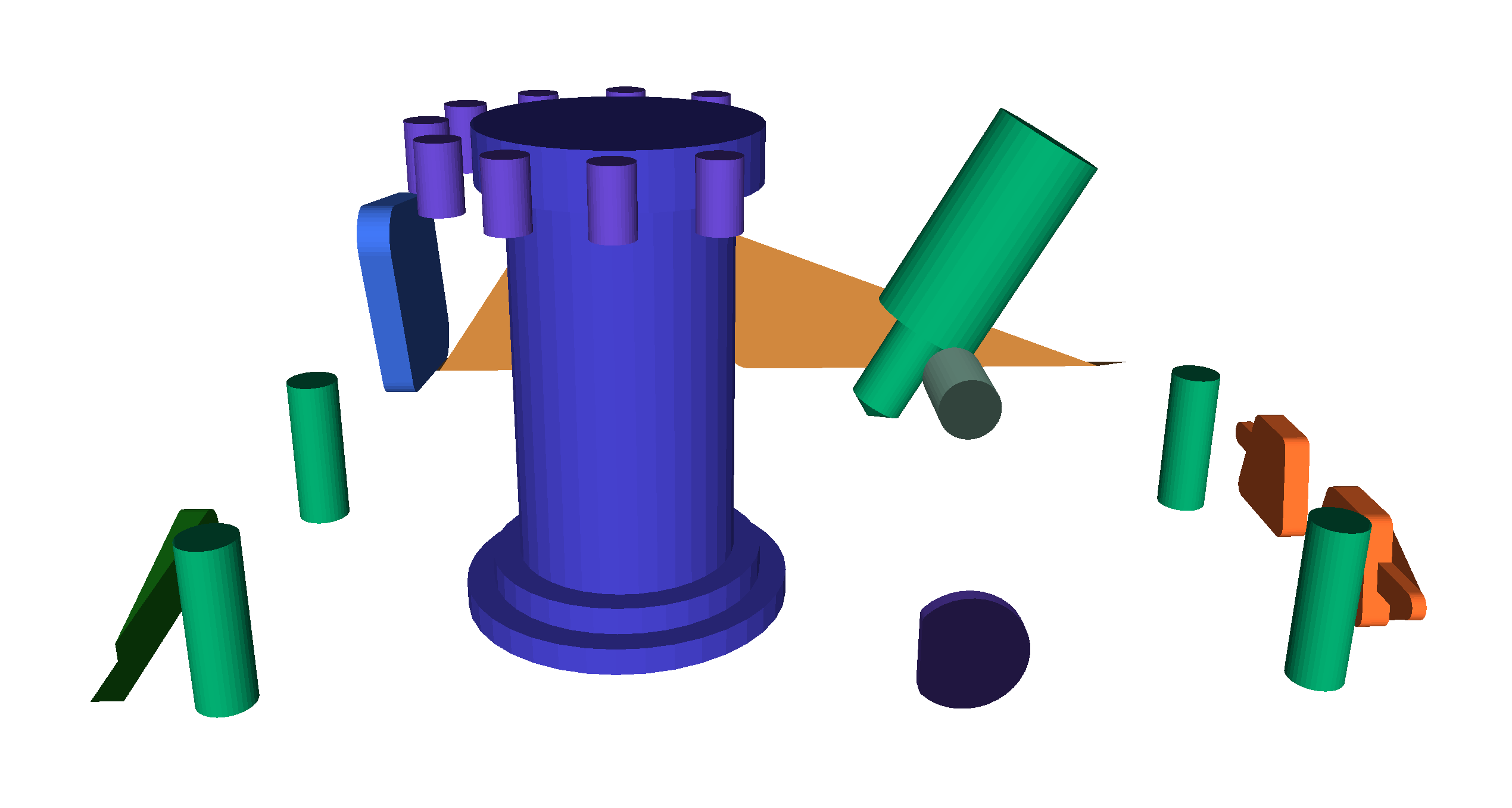}}\qquad \qquad
        \subfigure[\label{fig:443_in} CAD DEMO input.]
        {\includegraphics[width=.2\linewidth]{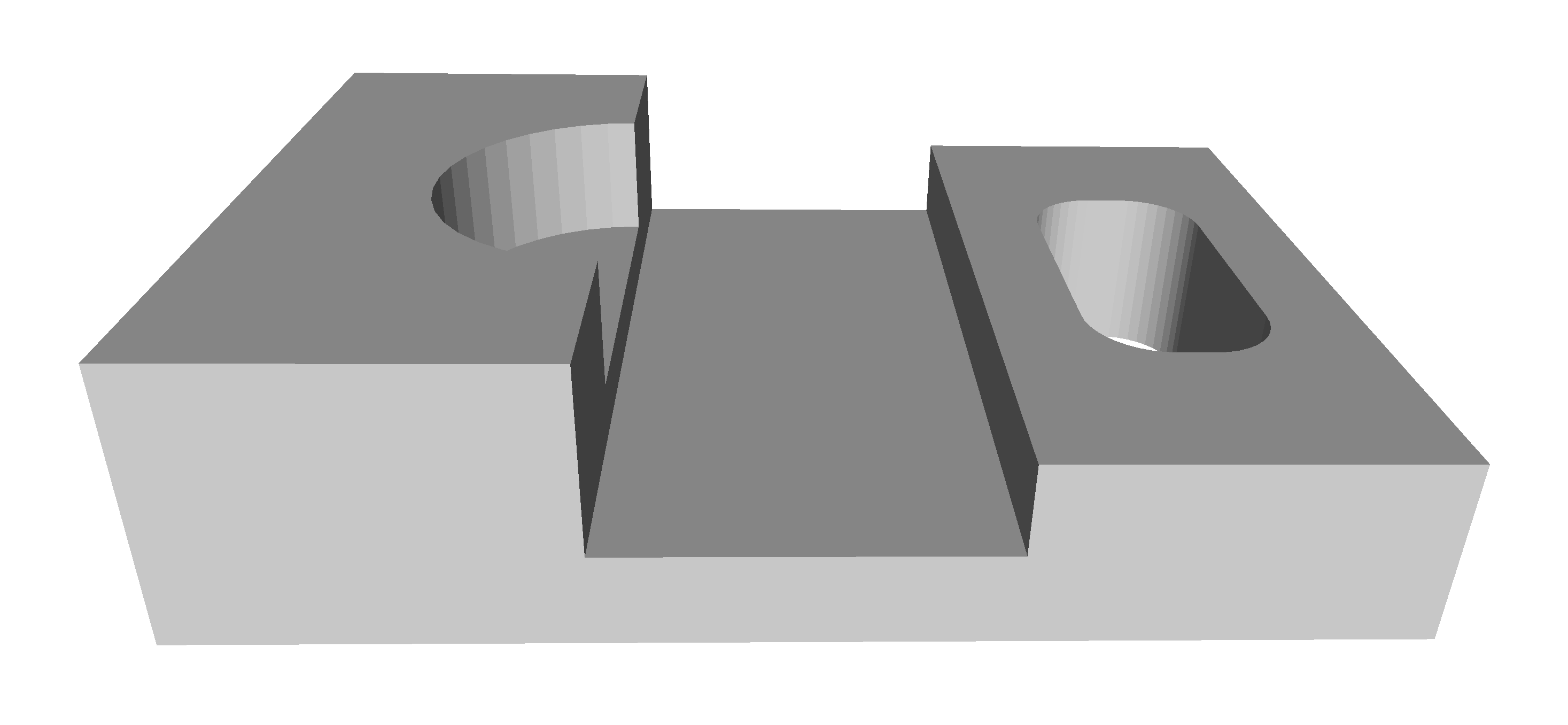}}
        \subfigure[\label{fig:443_out} CAD DEMO output.]
        {\includegraphics[width=.2\linewidth]{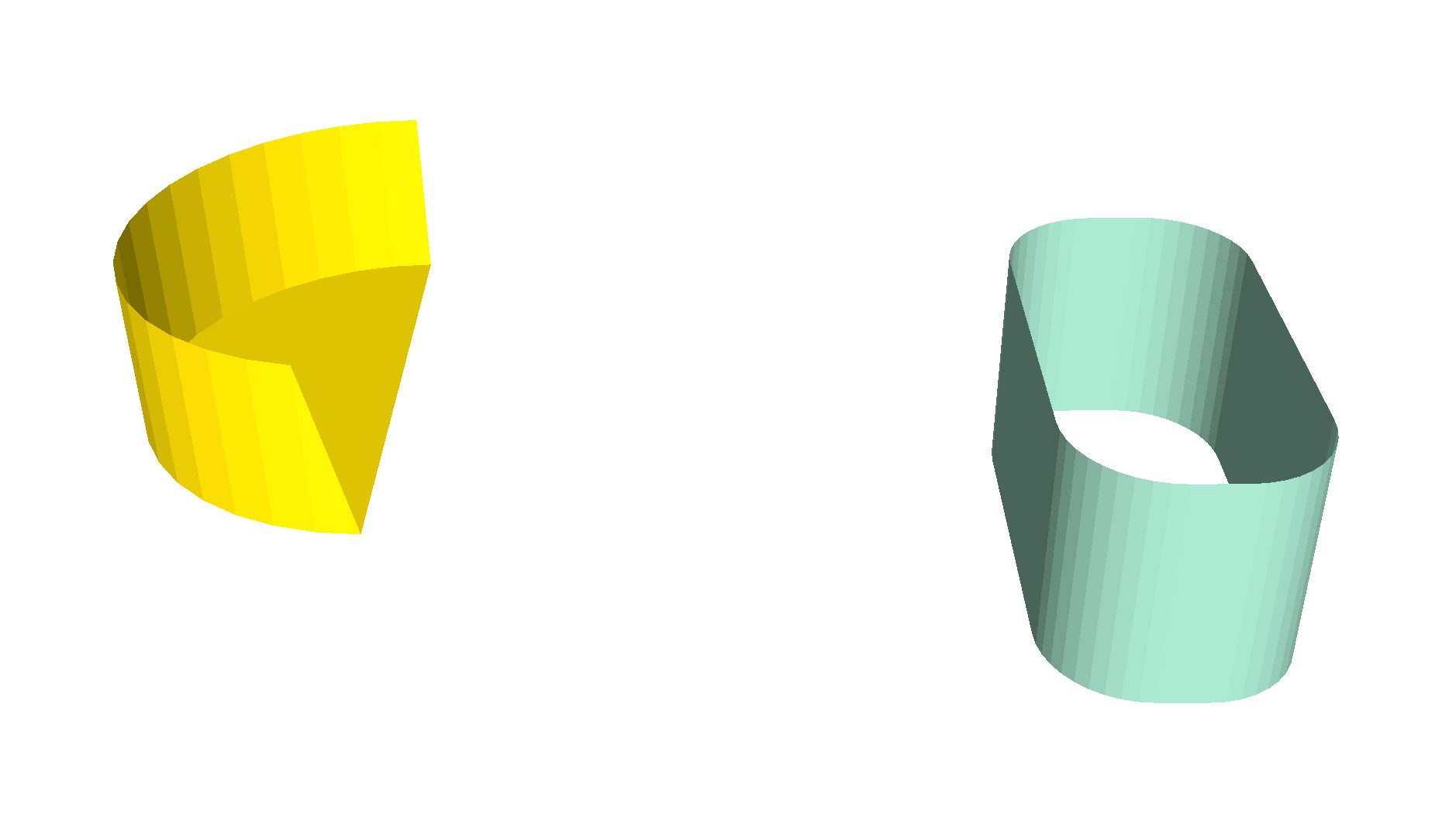}}\\
        
        \subfigure[\label{fig:840_in} team2 input.]
        {\includegraphics[width=.15\linewidth]{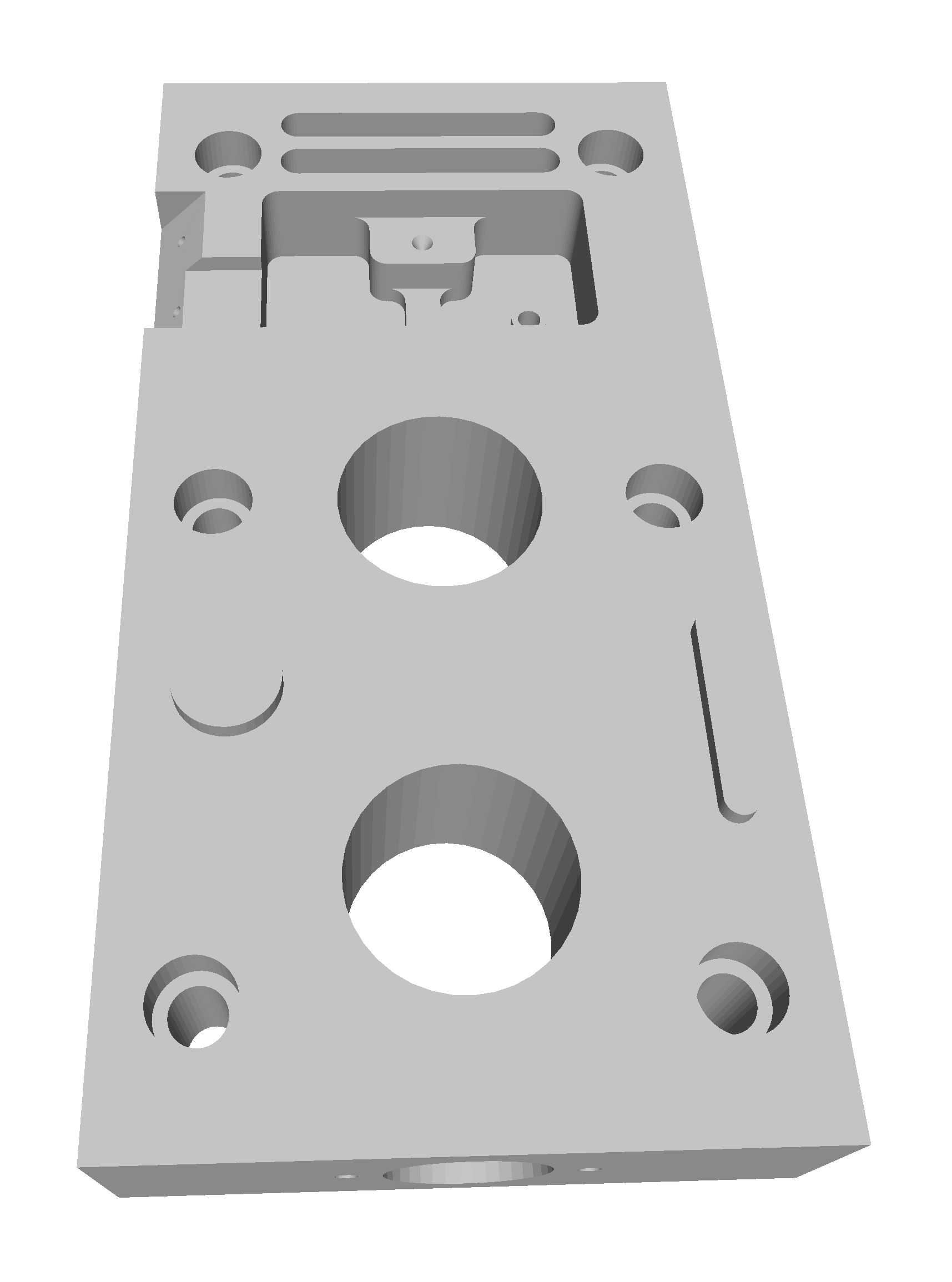}}
        \subfigure[\label{fig:840_out} team2 features.]
        {\includegraphics[width=.15\linewidth]{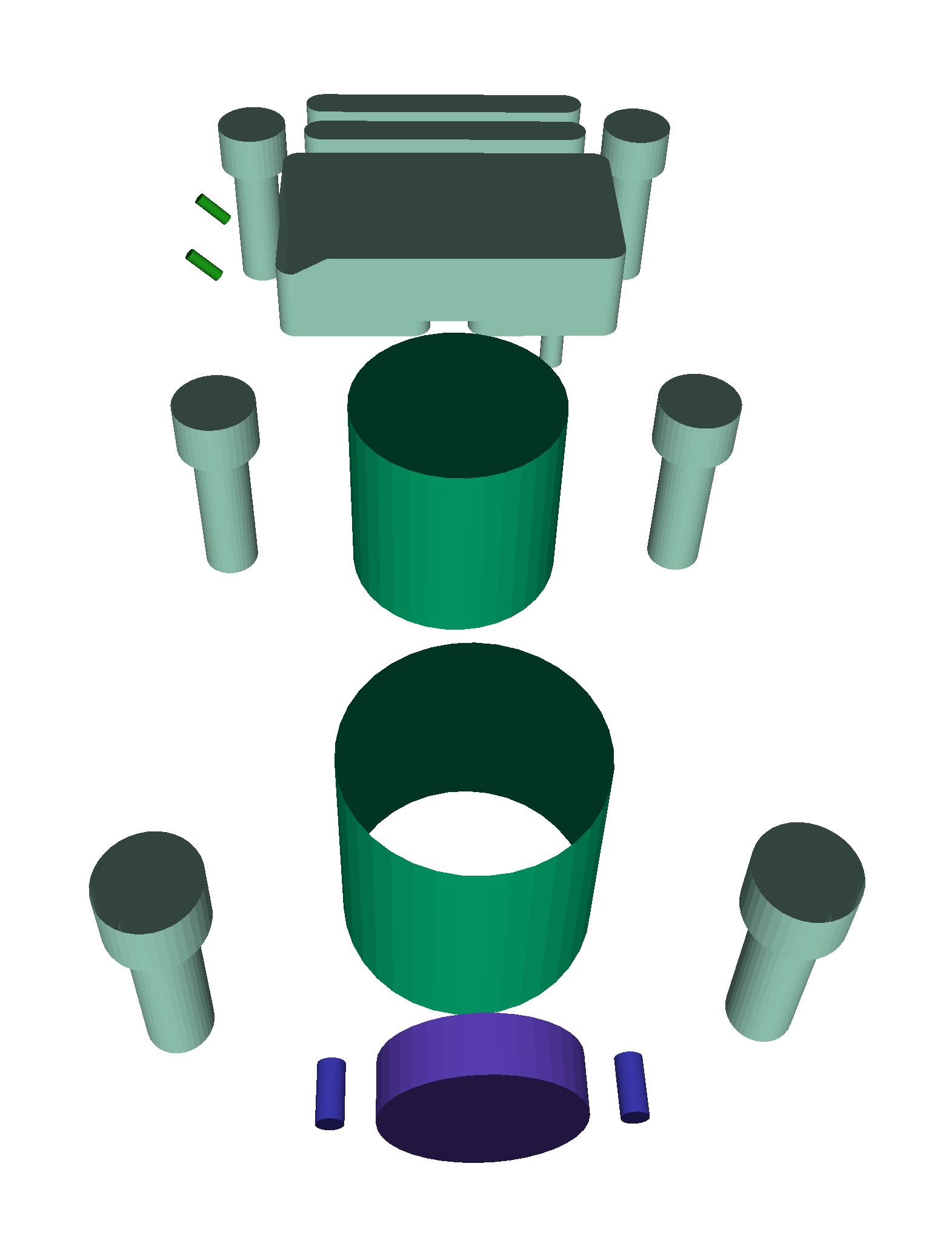}}\qquad \qquad
        \subfigure[\label{fig:benchmark_in} benchmark input.]
        {\includegraphics[width=.2\linewidth]{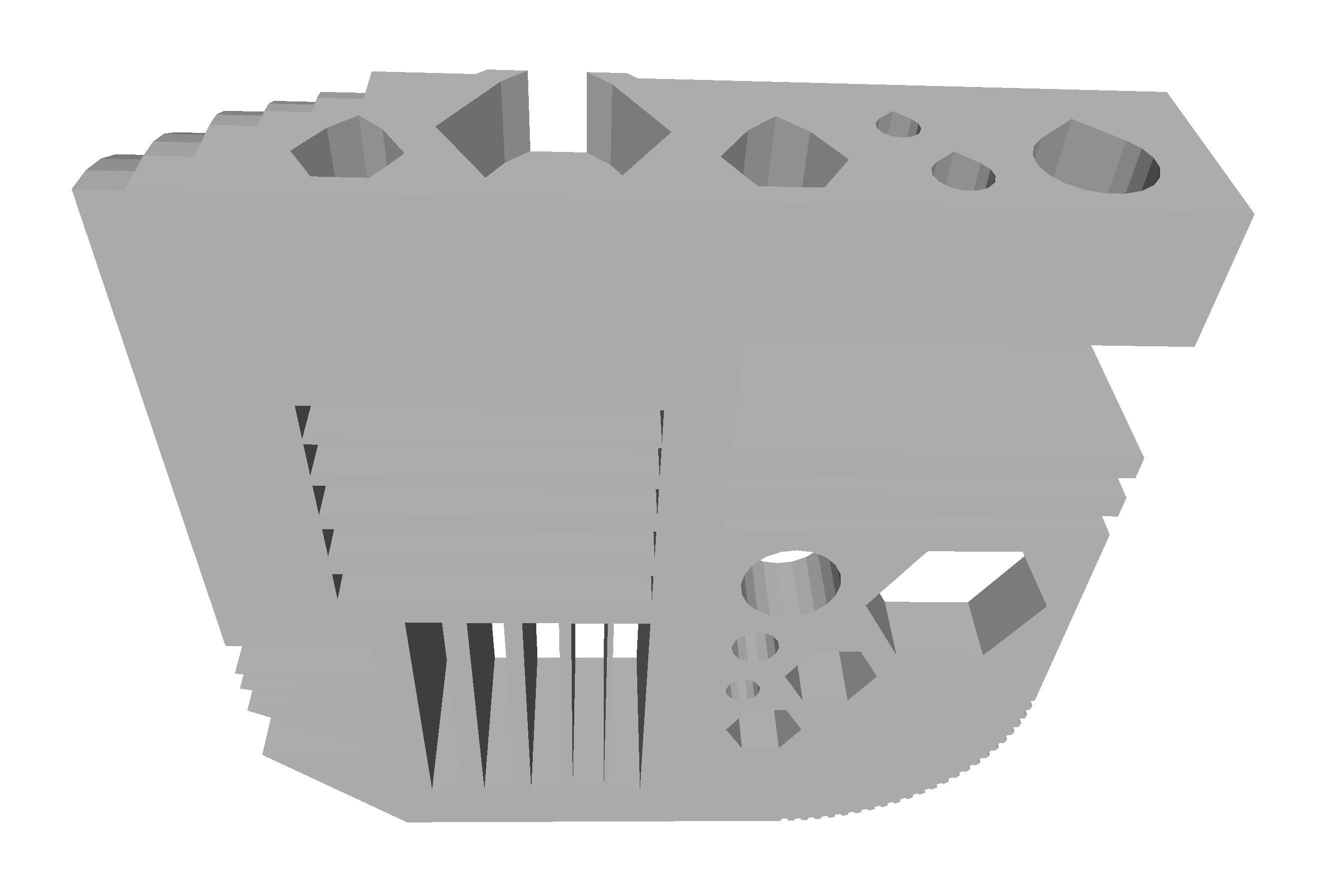}}
        \subfigure[\label{fig:benchmark_out} benchmark output.]
        {\includegraphics[width=.2\linewidth]{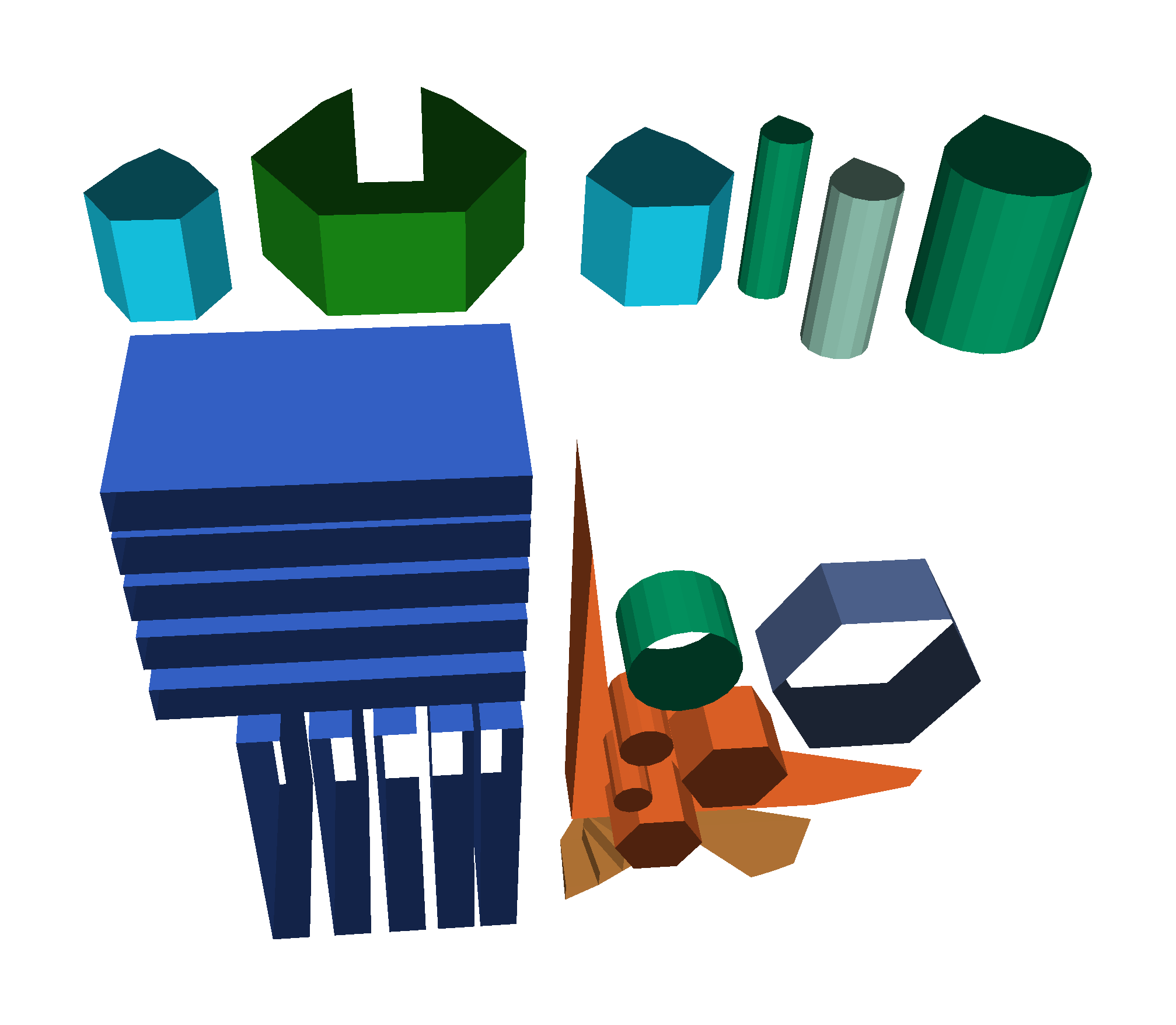}}\\
        
        \subfigure[\label{fig:holes10_in} holes10 input.]
        {\includegraphics[width=.12\linewidth]{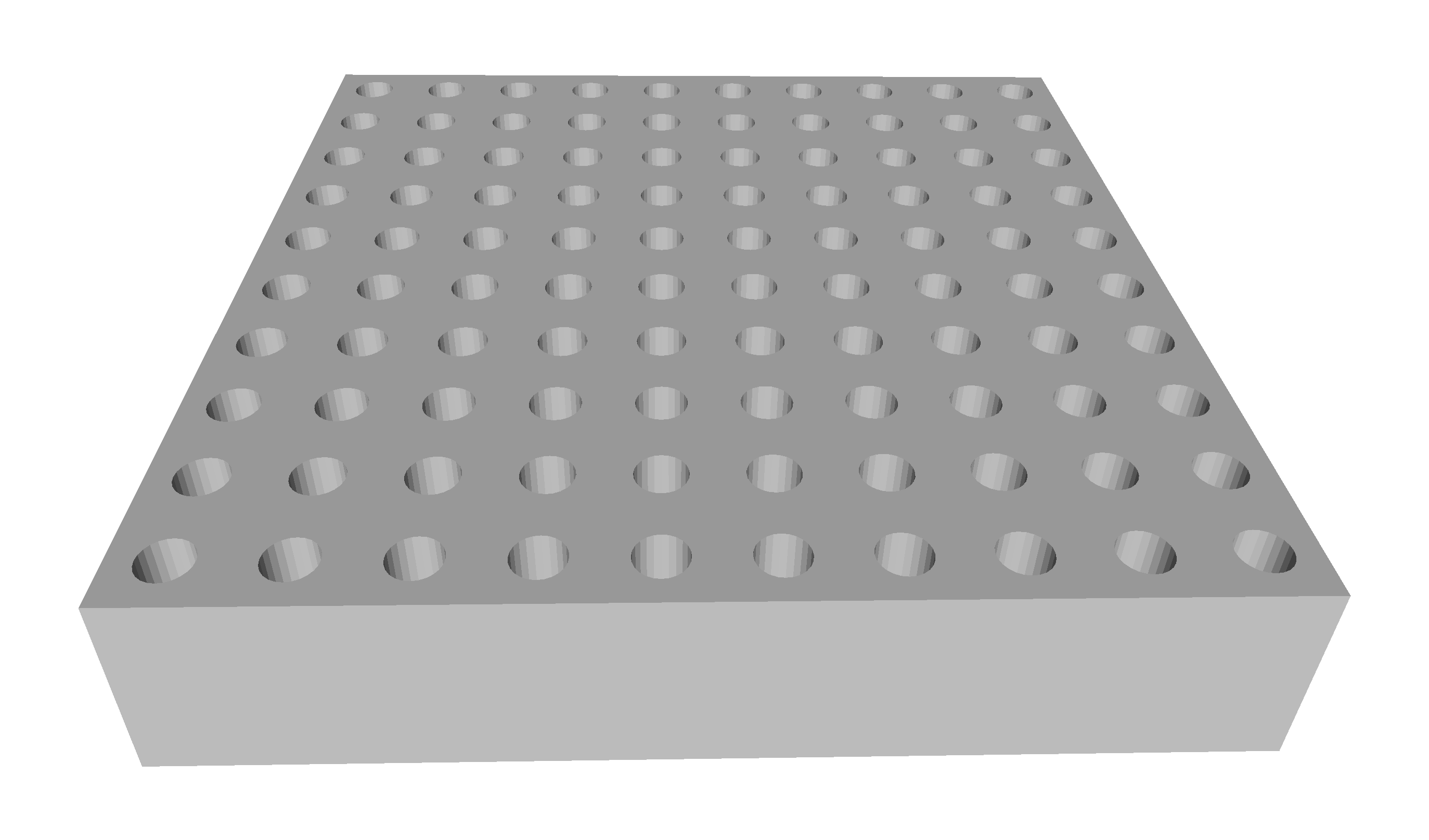}}
        \subfigure[\label{fig:holes10_out} holes10 output.]
        {\includegraphics[width=.12\linewidth]{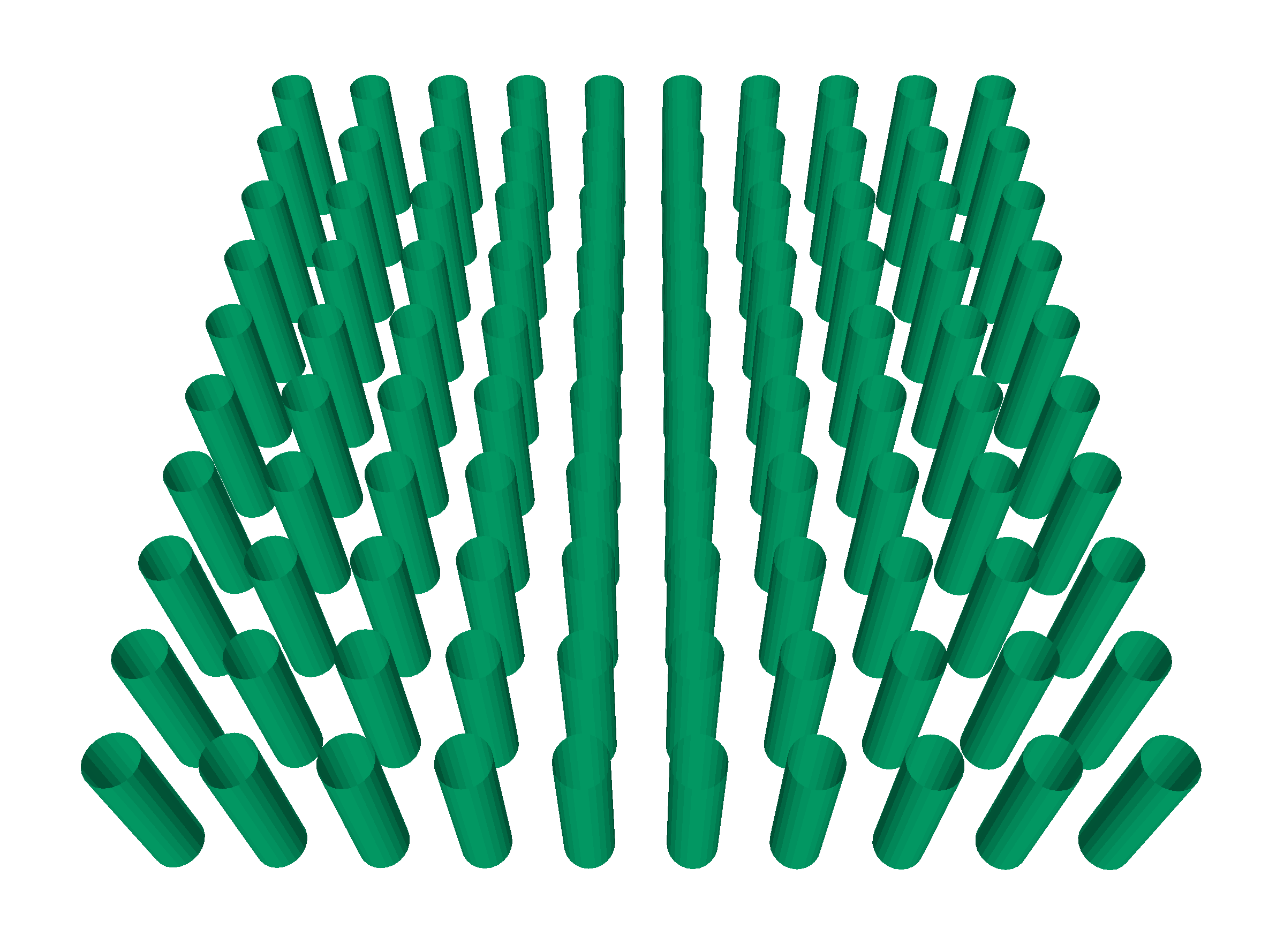}}
        \subfigure[\label{fig:gear38_in} gear38 input.]
        {\includegraphics[width=.12\linewidth]{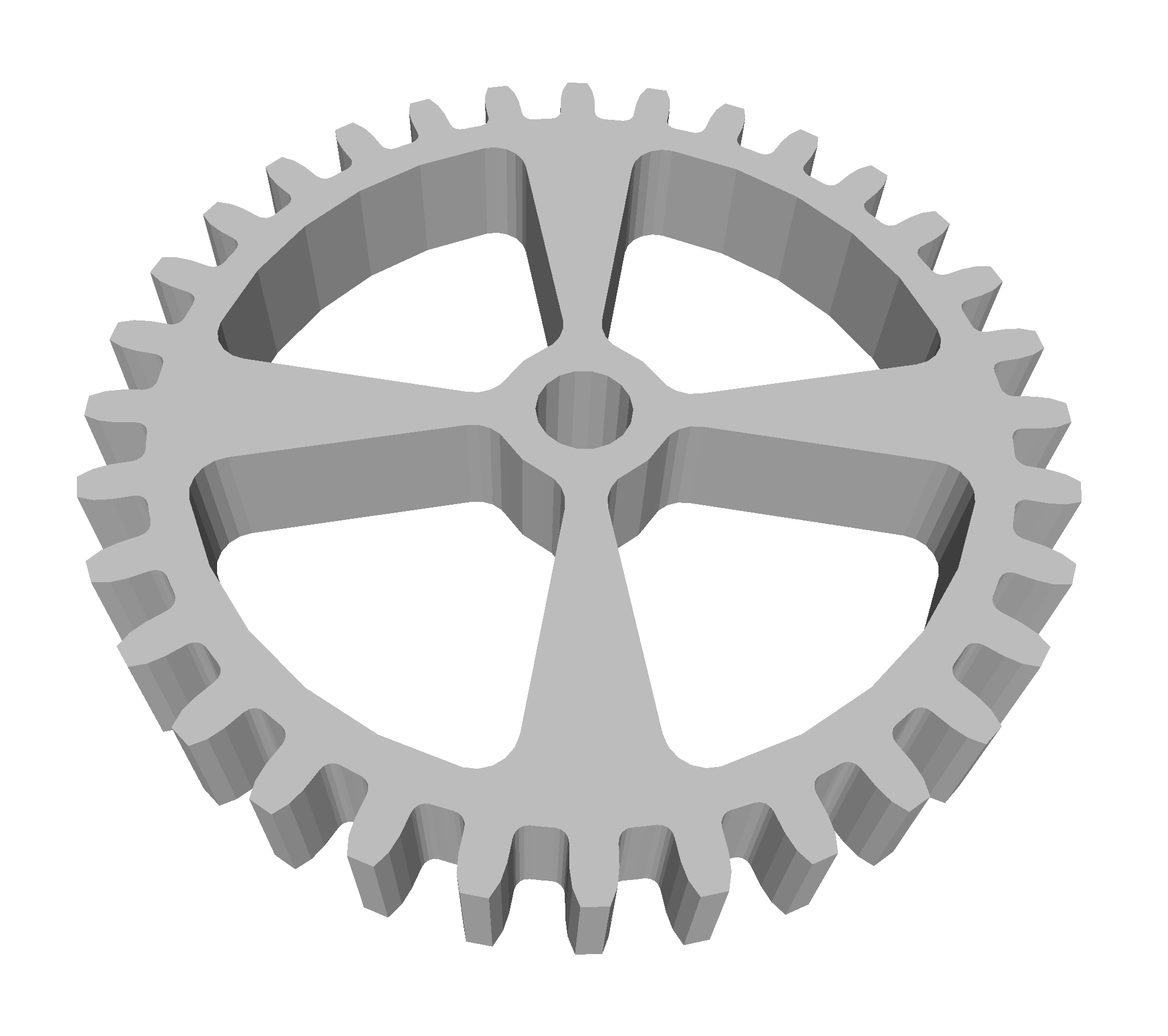}}
        \subfigure[\label{fig:gear38_out} gear38 output.]
        {\includegraphics[width=.12\linewidth]{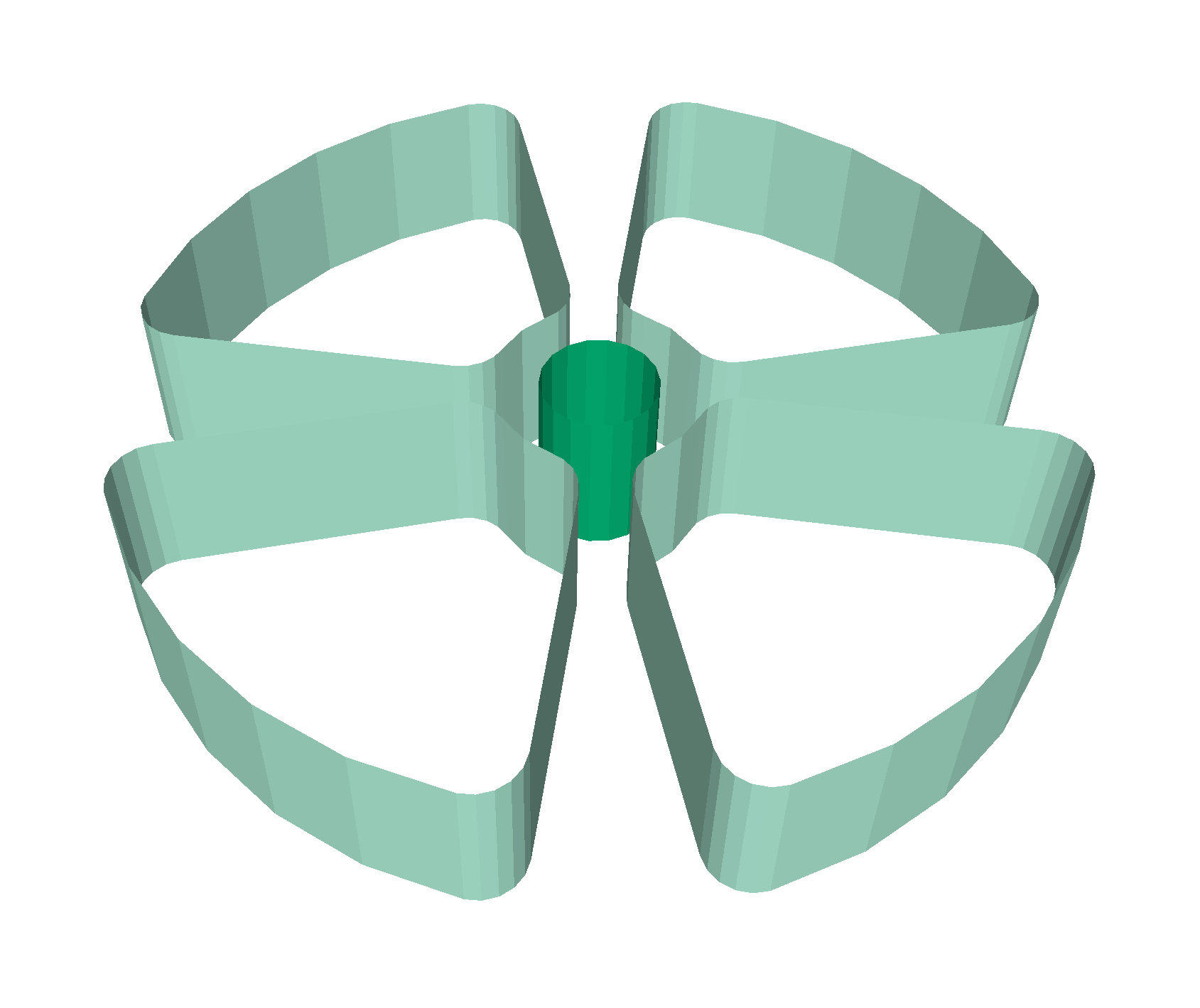}}
        \subfigure[\label{fig:442_in} 442 input.]
        {\includegraphics[width=.12\linewidth]{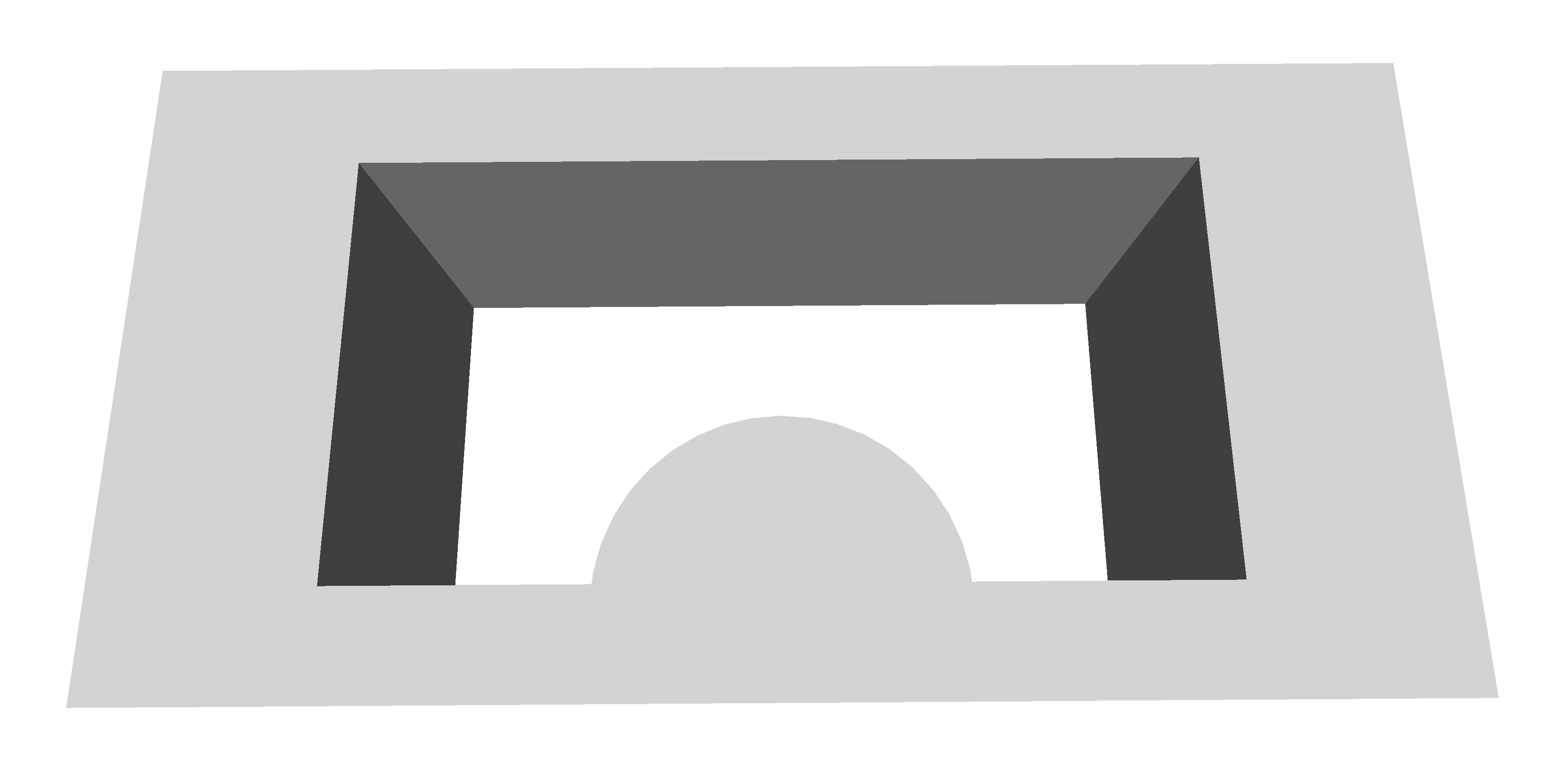}}
        \subfigure[\label{fig:442_out} 442 output.]
        {\includegraphics[width=.12\linewidth]{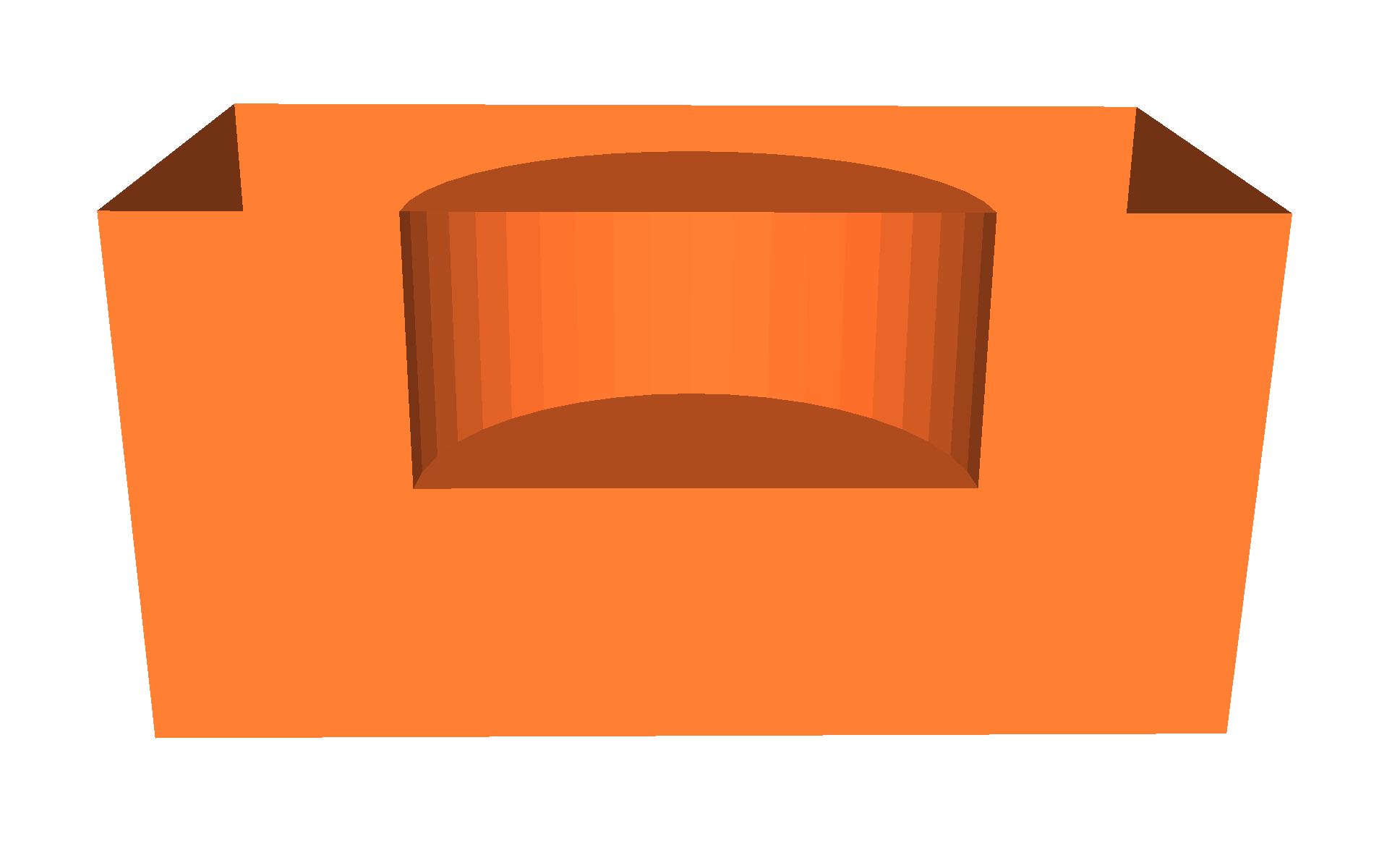}}\\
        
        \subfigure[\label{fig:base21_in} base21 input.]
        {\includegraphics[width=.12\linewidth]{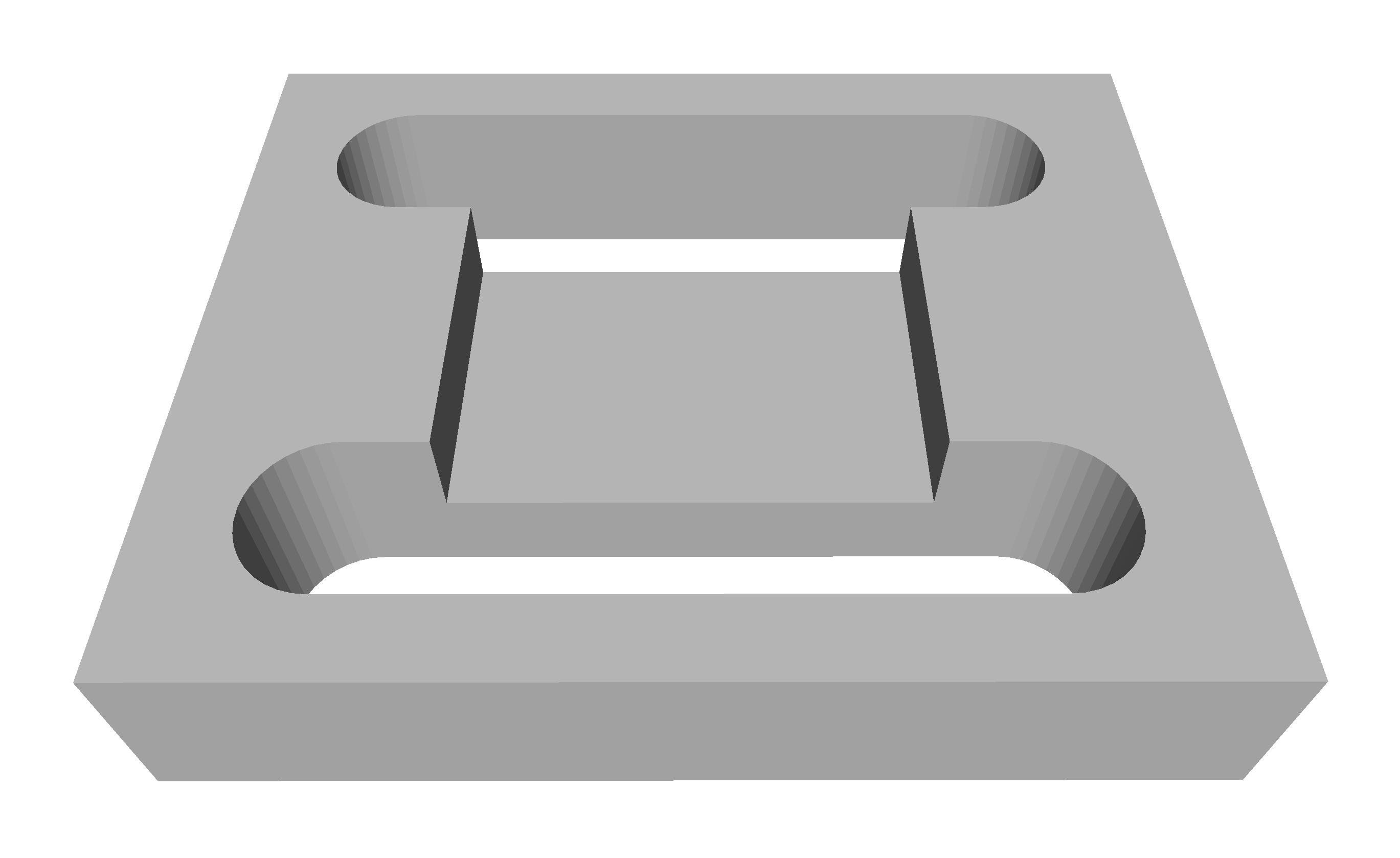}}
        \subfigure[\label{fig:base21_out} base21 output.]
        {\includegraphics[width=.12\linewidth]{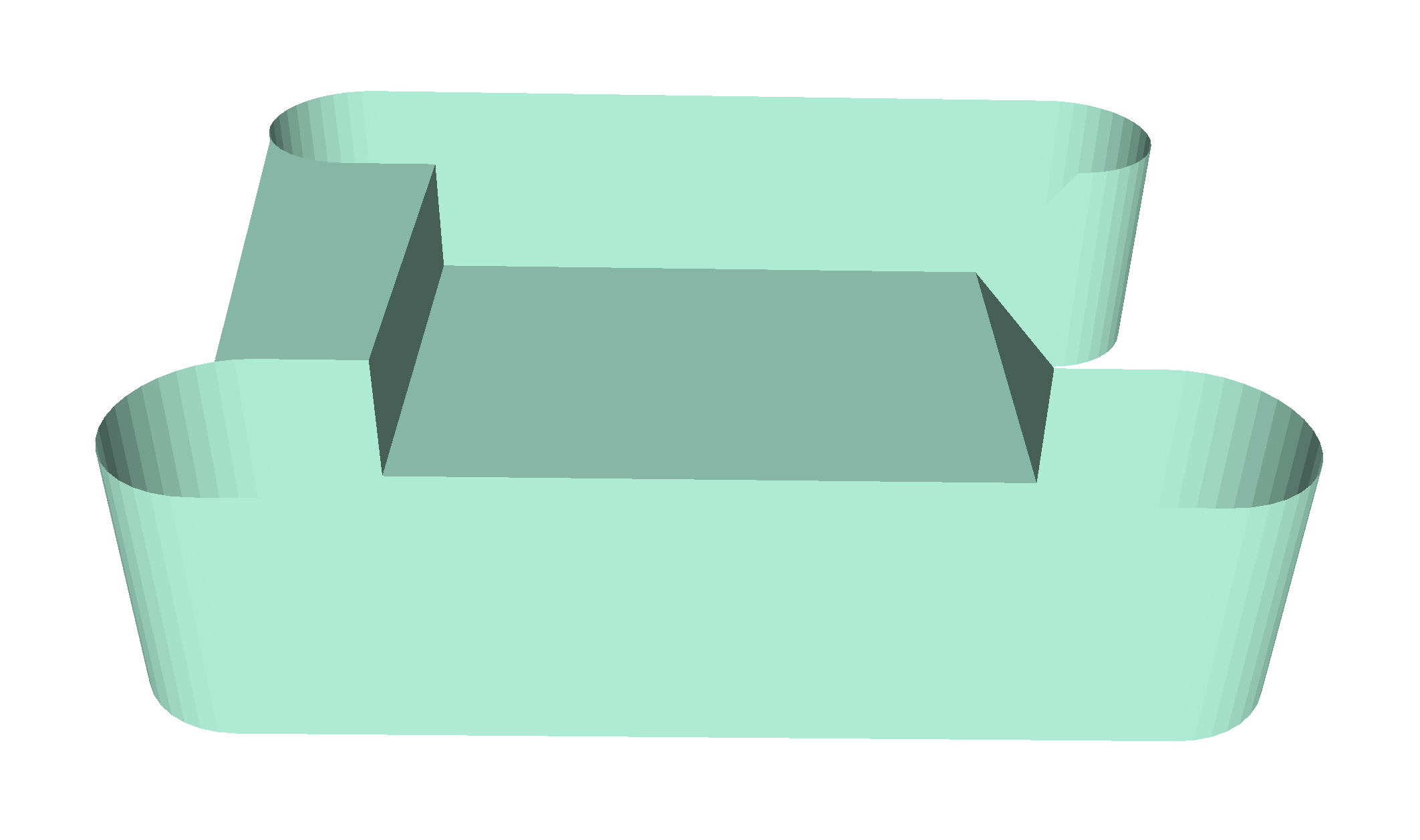}} 
        \subfigure[\label{fig:754_in} Screw input.]
        {\includegraphics[width=.12\linewidth]{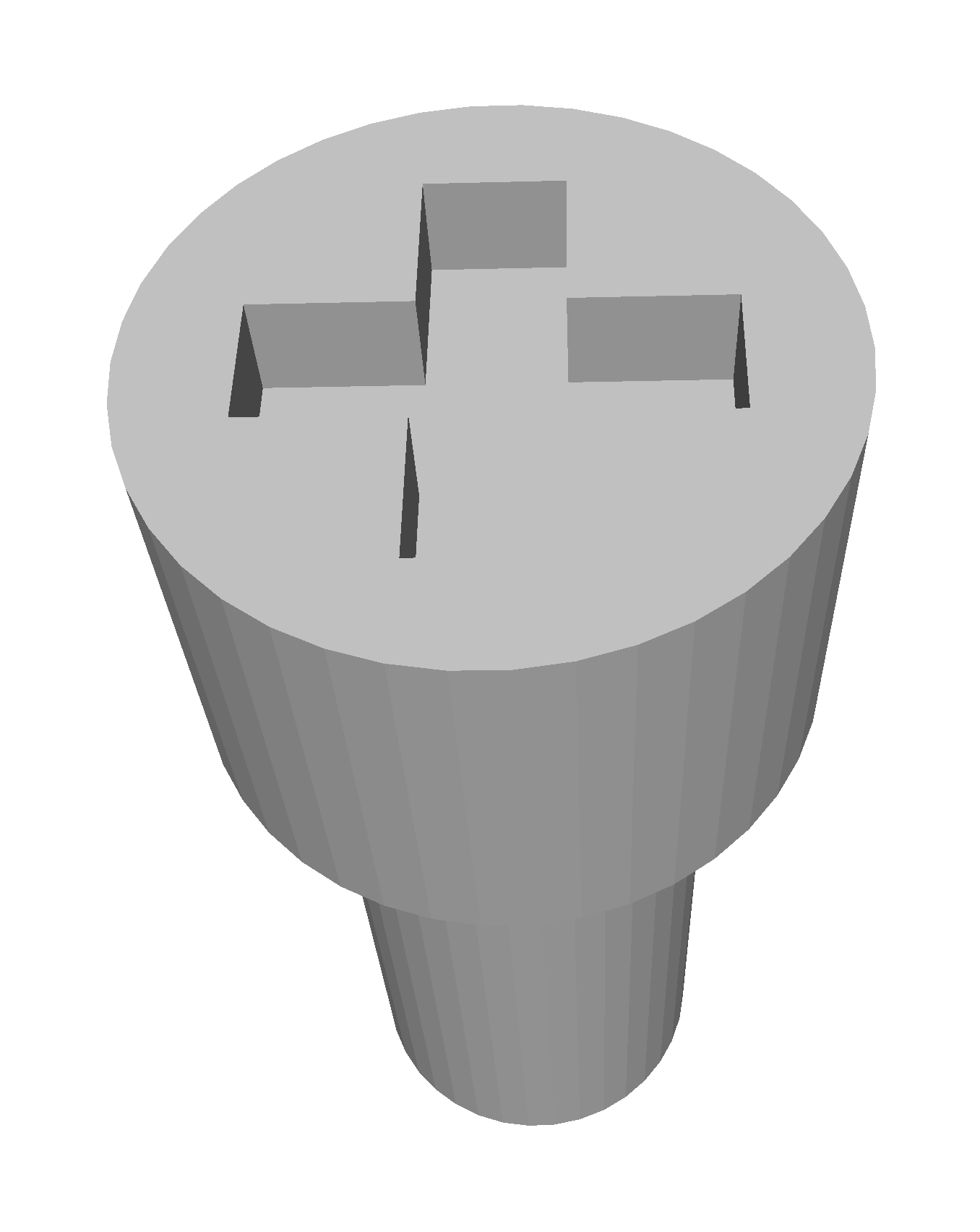}}
        \subfigure[\label{fig:754_out} Screw output.]
        {\includegraphics[width=.15\linewidth]{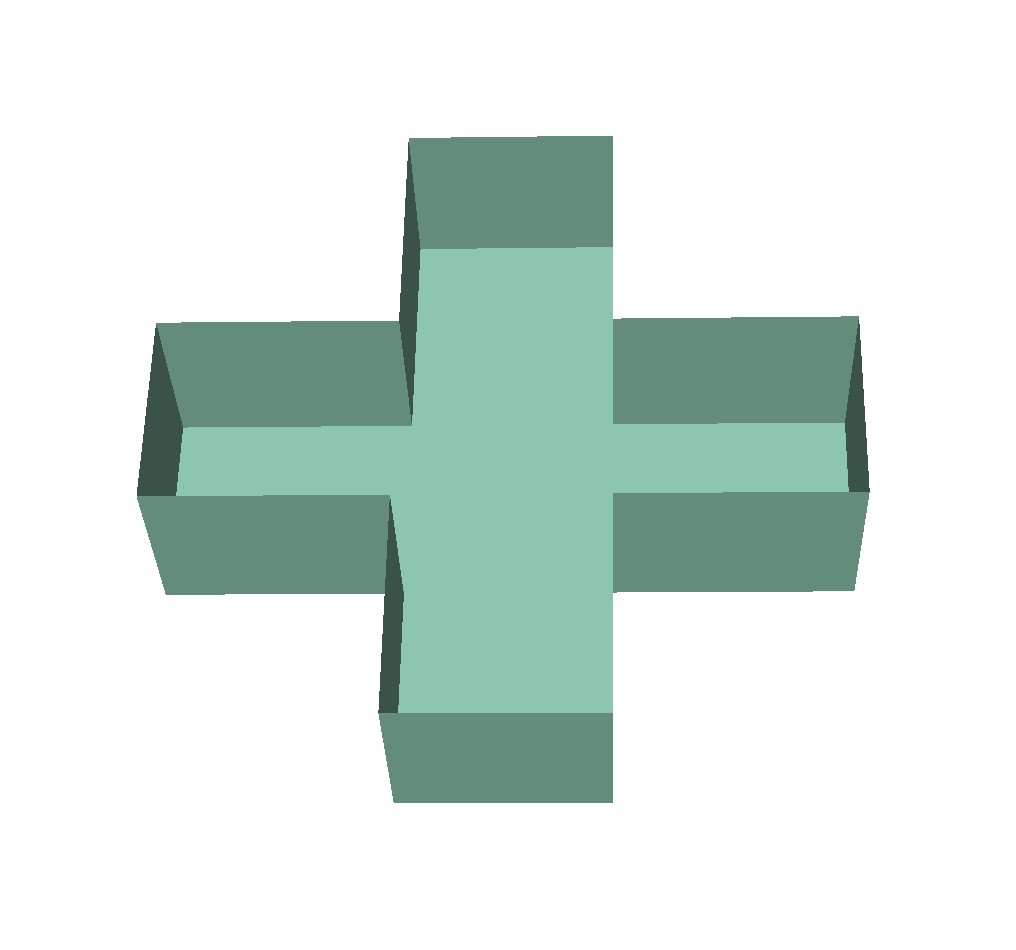}} 
        \subfigure[\label{fig:845_in} Screw input.]
        {\includegraphics[width=.15\linewidth]{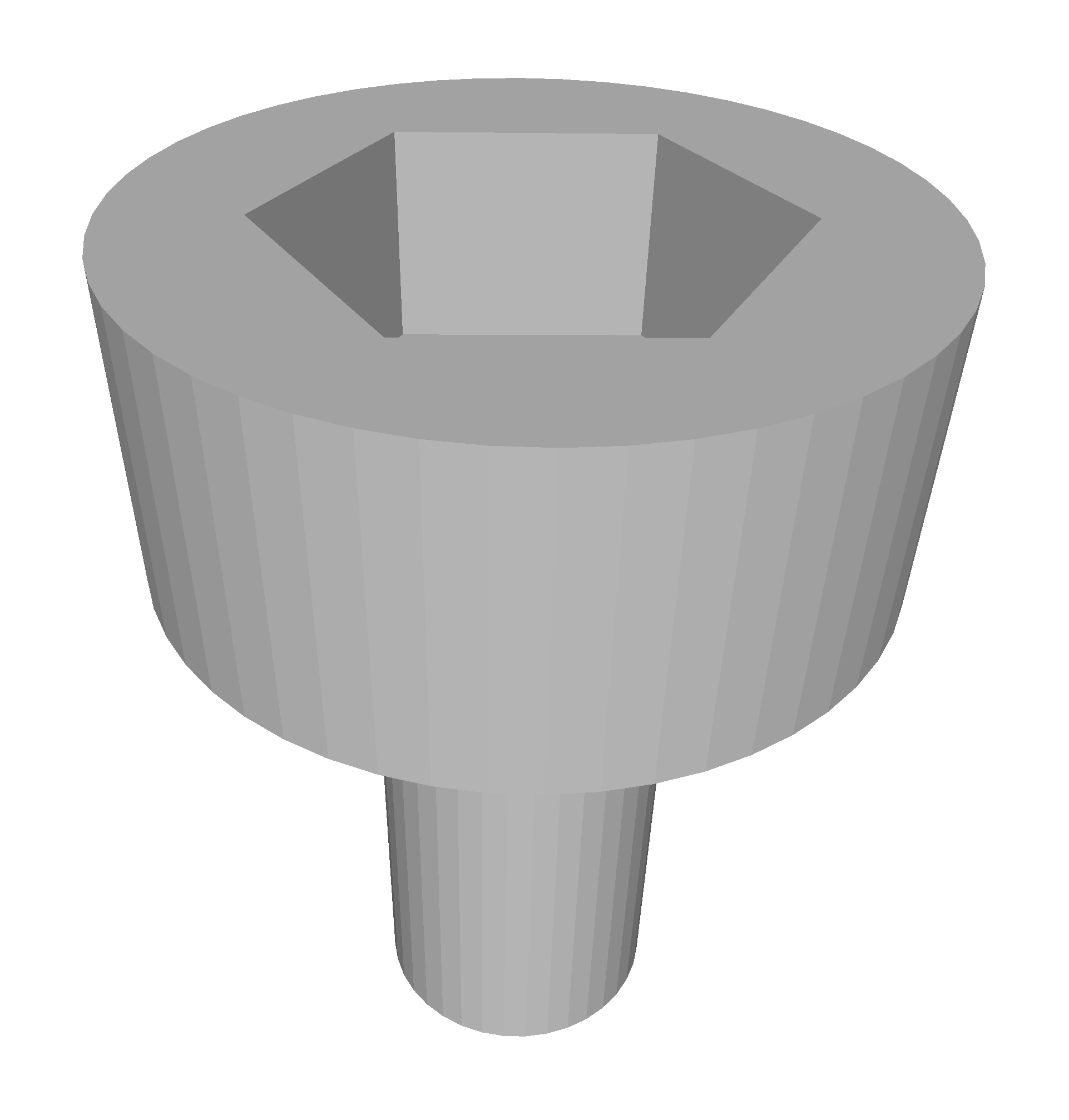}}
        \subfigure[\label{fig:845_out} Screw feature.]
        {\includegraphics[width=.15\linewidth]{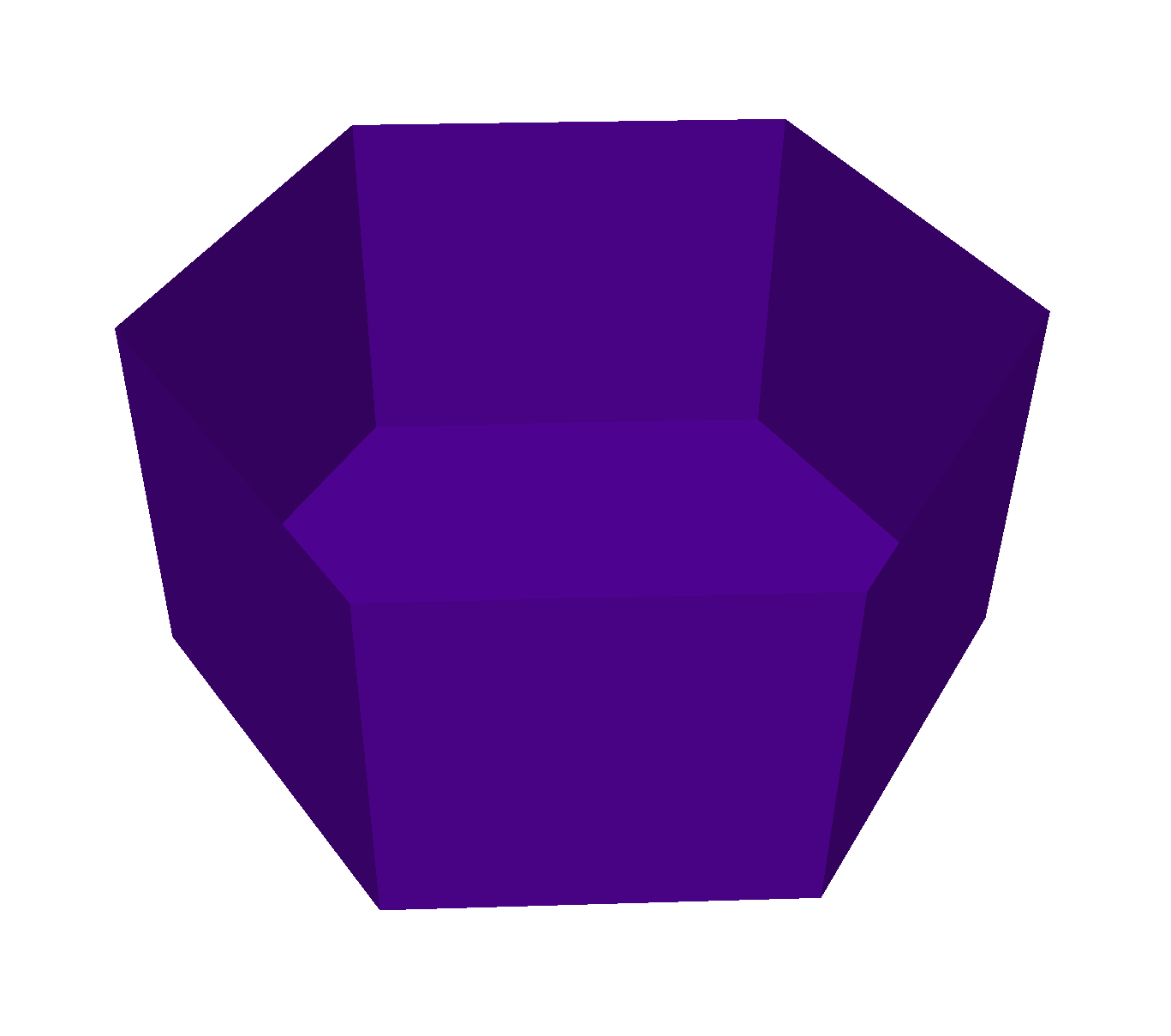}}
        \caption{ \label{fig:multi_feat} Results on models having multiple features. The models are obtained from ESB and NSR. The color of the features indicates that they have been recognised correctly when compared to single feature recognition (also see Figure \ref{fig:featuresingle}). A few of the features are complex/interacting and our approach is able to recognise them.}
    \end{minipage}
 \end{figure*}

We have experimented with various resolutions of Gauss map and came up with the resolution which works best for the feature classification. We have tried a reduced representation of 27 vertices and a higher representation of 227 vertices. For the reduced representation, the classification accuracy reduced to 85.15\% with a running time of 4.27 seconds and for a higher Gauss resolution, the classification accuracy remained at 86.20\% with a running time of 7.18 seconds. Therefore we can verify that reducing or increasing the resolution of the Gauss map will have an adverse affect on the performance of our feature classifier. The number of vertices on the Gauss map was chosen based on the shape distribution of the CAD models leading to a better classification accuracy.

\begin{table}[!h]
\begin{minipage}{\linewidth}
\centering
\begin{tabular}{|p{2.7cm}|p{2.0cm}|p{2.0cm}|}
\hline
Classifier  & Test Accuracy & Running Time (s)  \\  \hline
SVM (Sigmoid) & 31.85\% & 45.26s \\  \hline
SVM (RBF) & 40.98\% & 39.28s \\ \hline
SVM (Linear) & 76.81\% & 8.90s \\ \hline
Decision Tree & 95.59\% & 0.332s \\ \hline
XGBoost & 95.55\% & 41.89s\\ \hline
Random Forest & 97.90\% & 3.19s \\ \hline
\end{tabular}
\caption{\label{table:other_ml} Comparison with other machine learning algorithms}
\end{minipage}
\begin{minipage}{\linewidth}
\centering
\begin{tabular}{|p{2.3cm}|p{2.5cm}|p{2.0cm}|}
\hline
 & FeatureNet  & Our Method \\ \hline
Training Accuracy & 99\% & 100\% \\  \hline
Testing Accuracy & 97.4\% & 97.90\% \\  \hline
Run Time & 23400s & 3.19s \\  \hline
Hyperparameters & Convolutional layers=4, voxel size=64x64x64 & no of tree=130 , depth=100 \\ \hline
No of models for 24 classes & 144000 & 24000 \\ \hline
\end{tabular}
\caption{Comparison with FeatureNet  \cite{ZHANG201812}}
\label{table:FeatureNet_compare}
\end{minipage}
\end{table}

\subsubsection{Comparison with other recognition algorithms} \label{sec_compclassifiers}
We have tried other classical machine learning algorithms such as SVM \cite{Hearst:1998:SVM:630302.630387}, Decision tree \cite{Breiman:2253780}, XGBoost \cite{Chen:2016:XST:2939672.2939785} etc and the results are summarized in Table \ref{table:other_ml}. The parameters were tuned to find the best performances. Random Forest method has performed the best among all the other methods with the highest accuracy and also has taken much lesser running time (except the decision tree based approach).

\subsection{Multi feature recognition}

The proposed approached was then tested on multi-feature models from databases such as ESB \cite{Jayanti06} and NDR \cite{RegliNDS} which are also unseen as they were not included in the training set. The features are extracted first using the algorithm in \cite{MURALEEDHARAN201851} and recognised using the proposed approach. The results of the recognition for a selected subset are illustrated in Figure \ref{fig:multi_feat}. Color of the recognised features indicates the correct recognition when compared to single features as in Figure \ref{fig:featuresingle}.   

In order to perform multi-feature recognition, we followed a similar strategy as that of single feature recognition using the same feature extraction algorithm. Once all the features are extracted, each one of the features was aligned as explained in section \ref{sec_align} and its $[102\times1]$ shape signature was calculated as in section \ref{sec_Gauss}. Each of the $[102\times1]$ shape signature corresponds to one of the features existing in the model. Now using the pre-trained random forest model, each $[102\times1]$ vector is input to find the output label predicted by the model. From the class label, the predefined color was allotted to each of the features for visualizing the predicted class. 

From analyzing the results in Figure \ref{fig:multi_feat}, we can clearly see that all the features with categories which were trained were correctly recognised by the method. 
For e.g., for models in Figures \ref{fig:268_in}, \ref{fig:443_in}, \ref{fig:840_in}, etc., all features of trained categories are recognised correctly. Also, holes10 model in Figure \ref{fig:holes10_in} which contains 100 through holes are all recognised correctly. 

\subsection{Complex/interacting feature recognition}

In the ESB and NDR benchmark datasets, there are a lot of feature categories where a correct recognition could not be given. In such cases, the trained Random forest has assigned the most similar class as per the geometry of the object. For example, the gear model shown in Figure \ref{fig:gear38_in} has a curved hole feature which is recognised as a circular end through hole. Similarly, base21 model in Figure \ref{fig:base21_in}  contains two curved holes joined together, it is recognised as a circular end through hole. In Figure \ref{fig:754_in}, the '+' shape of the slot cannot be exactly categorised, but it can be thought of multiple rectangular slots joined to each other. The model has labeled the extracted feature as a rectangular slot as shown in Figure \ref{fig:754_out}.

\begin{figure}[!h]
	\centering
        \subfigure[\label{fig:1236922_in} 1236922 input.]
        {\includegraphics[width=.2\linewidth]{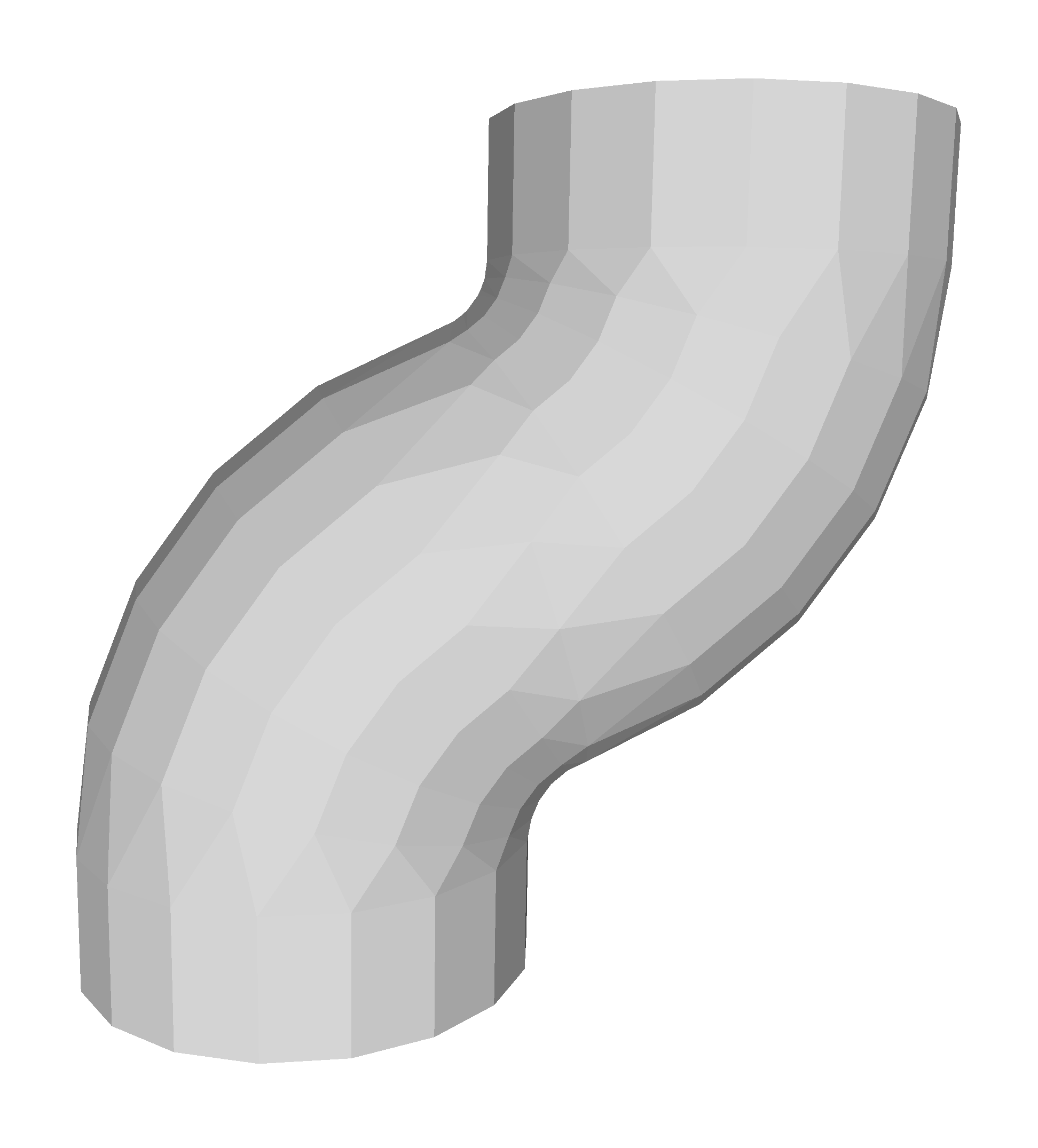}}
        \subfigure[\label{fig:1236922_out} 1236922 output.]
        {\includegraphics[width=.2\linewidth]{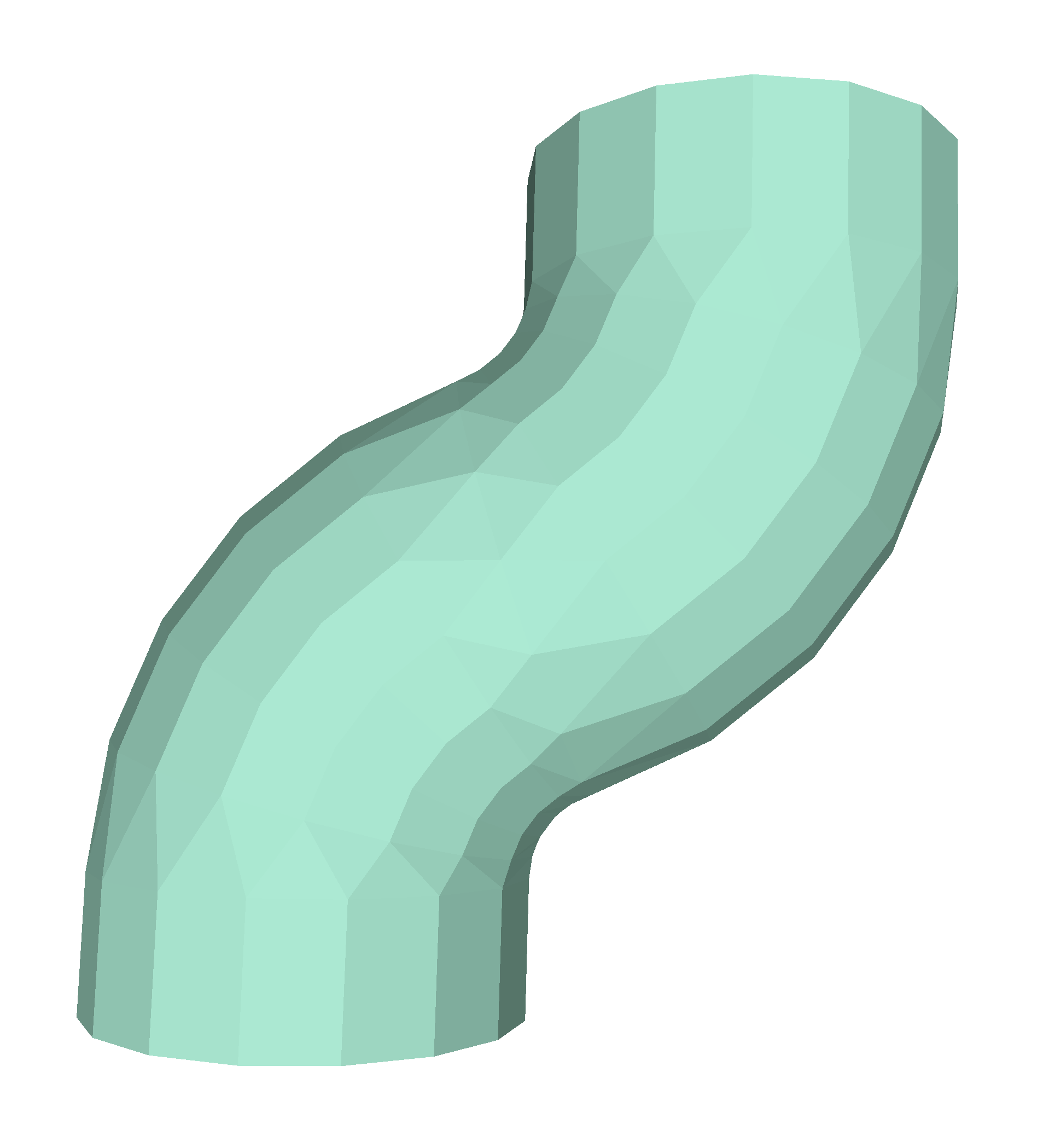}}
        \subfigure[\label{fig:1278555_in} 1278555 input.]
        {\includegraphics[width=.13\linewidth]{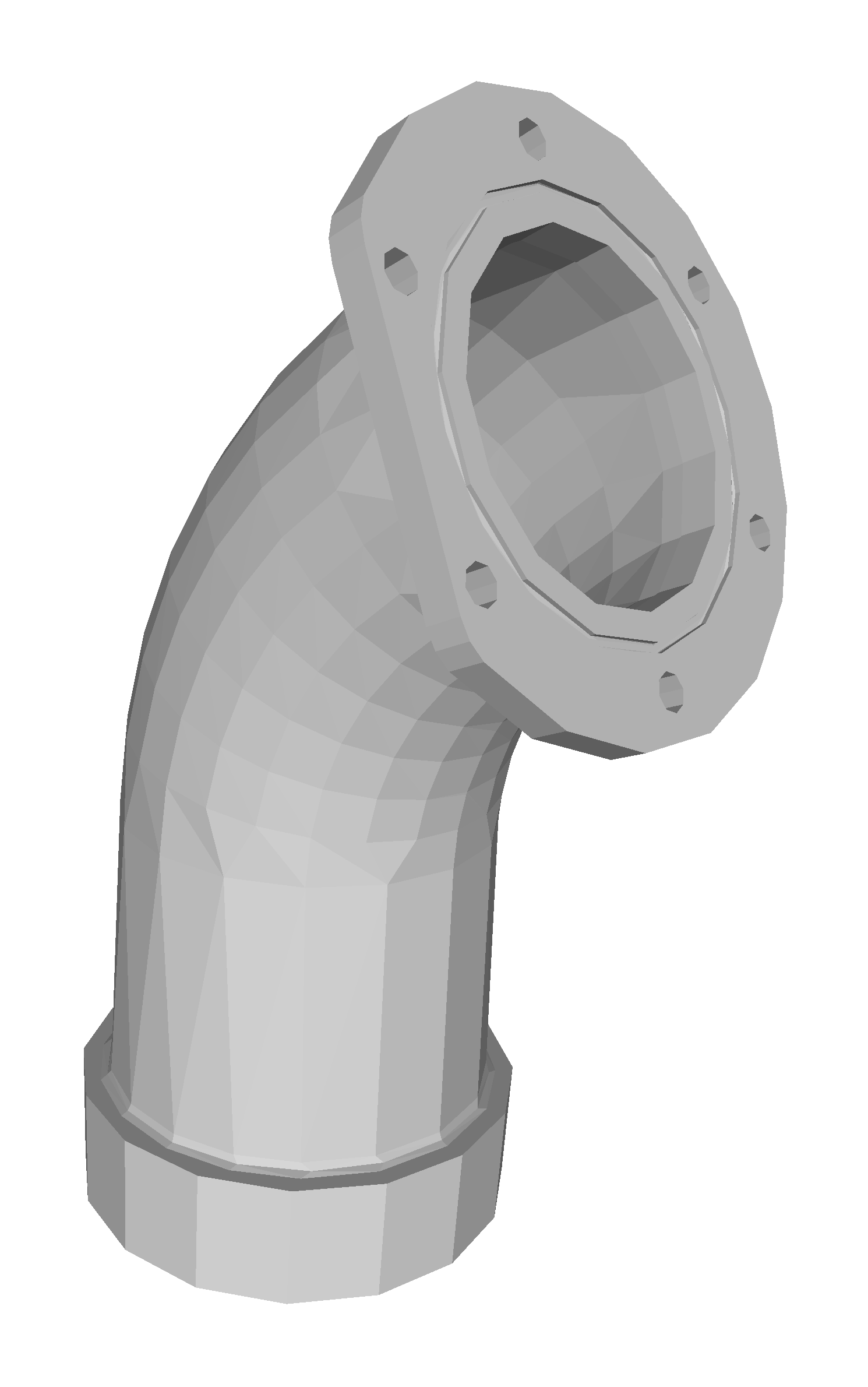}}
        \subfigure[\label{fig:1278555_out} 1278555 output.]
        {\includegraphics[width=.15\linewidth]{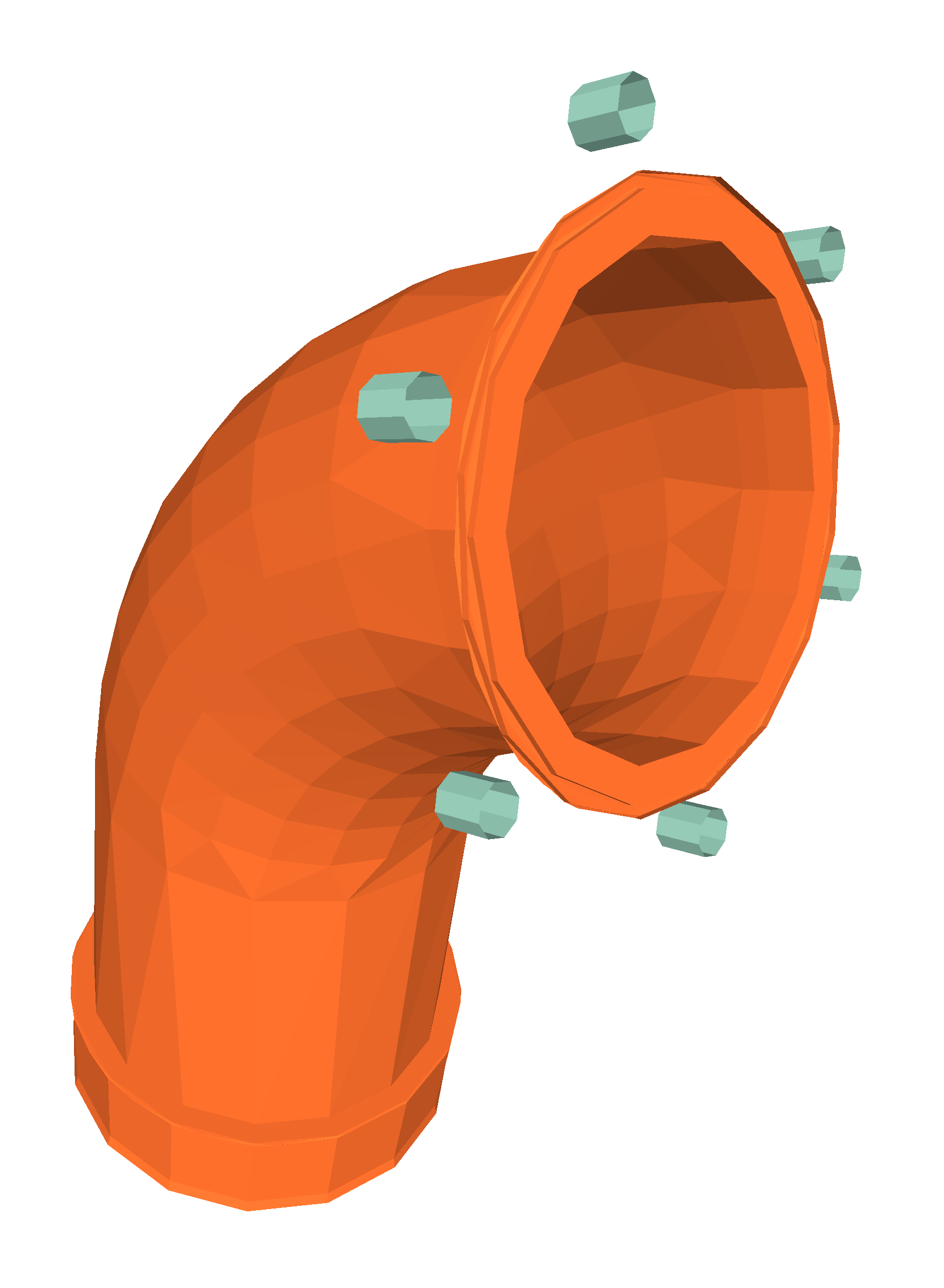}}
\caption{\label{fig:unseen} Unseen models from pipe category of ESB database}
\end{figure}

\subsection{Recognition of unseen features}

It is interesting to observe the behavior of the recognition algorithm on unseen data. Figure \ref{fig:unseen} shows the feature recognition of unseen category - pipe models. It shows two geometrically dissimilar pipe objects recognised differently. The model in Figure \ref{fig:1236922_in} is recognised as circular end through hole and the model in Figure \ref{fig:1278555_in} is recognised as circular blind slot.

\subsection{Recognition of features in a noisy data}

\begin{figure}[!h]
	\centering
	\subfigure[\label{fig:block_out} BBR=0]
	{\includegraphics[width=.45\linewidth]{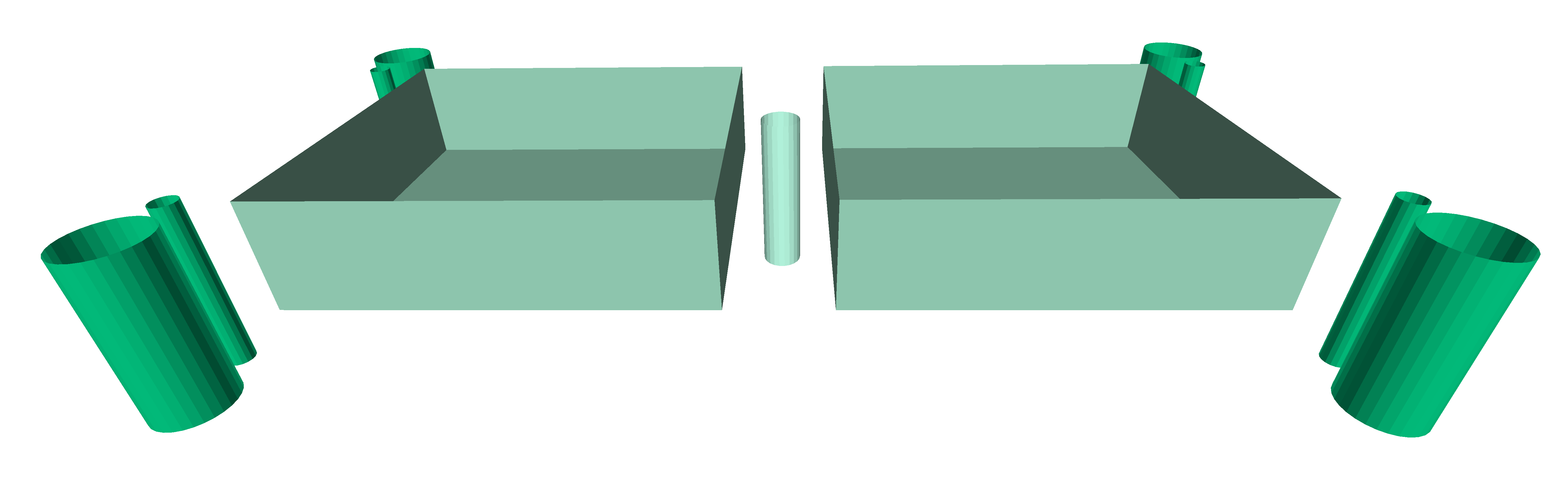}}
	\subfigure[\label{fig:block10_out} BBR=10 ]
	{\includegraphics[width=.45\linewidth]{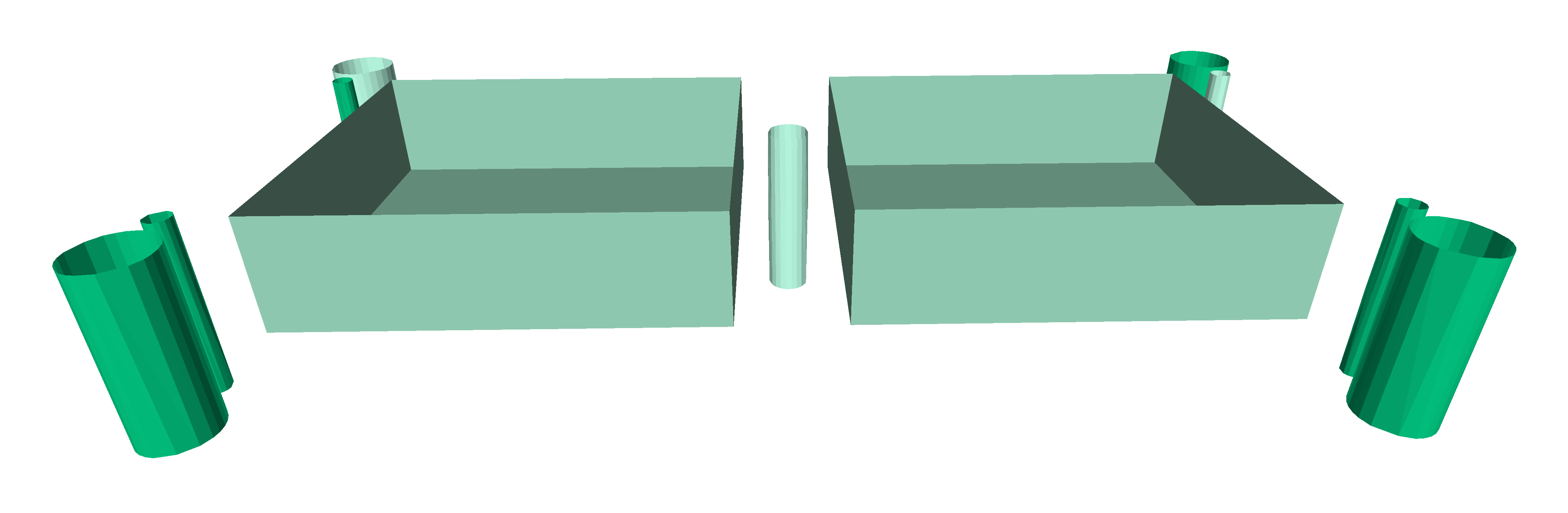}}
	\subfigure[\label{fig:block50_out} BBR=50 ]
	{\includegraphics[width=.45\linewidth]{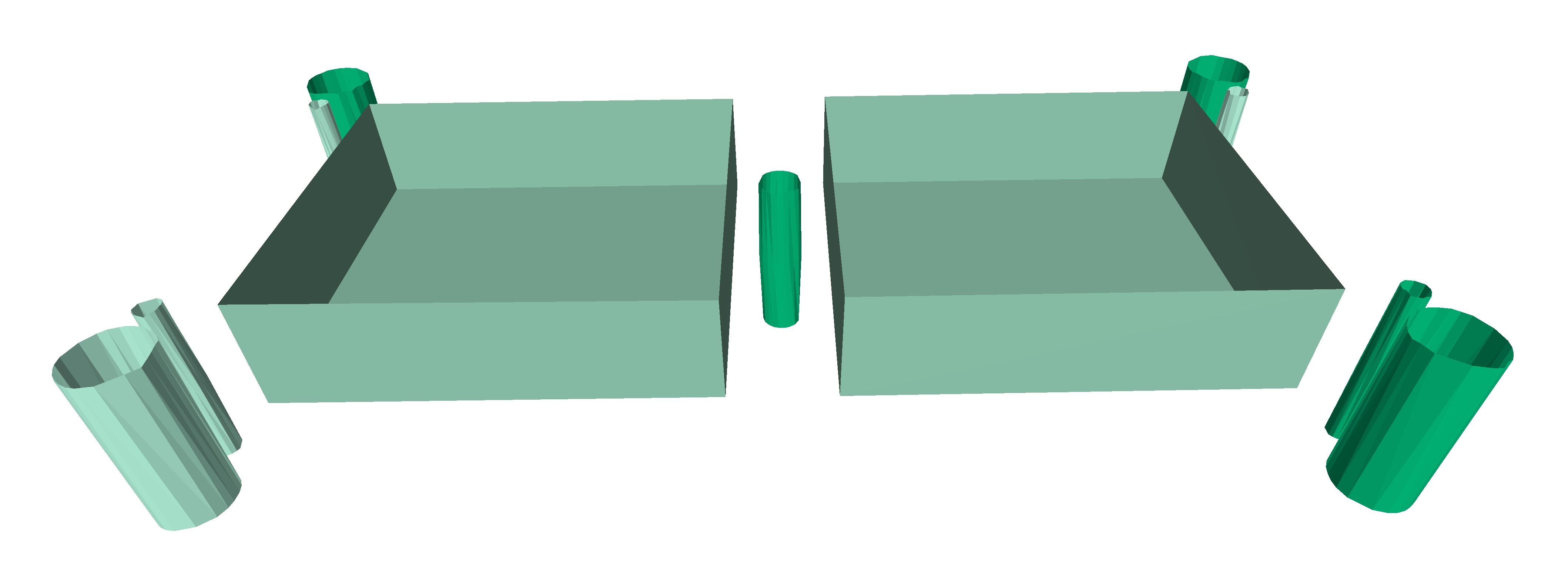}}
	\subfigure[\label{fig:block100_out} BBR=100]
	{\includegraphics[width=.45\linewidth]{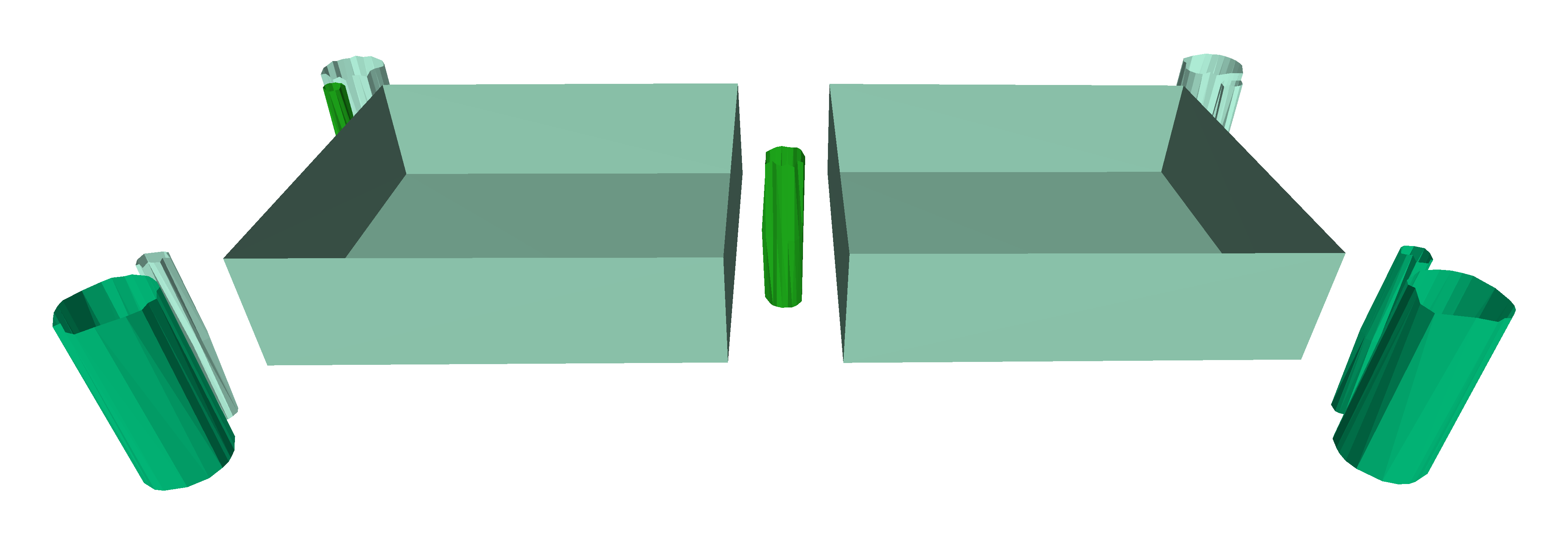}}
	\caption{\label{fig:noisy inputs}Recognition of features with noisy models with Bounding Ball Radius (using ReMesh software \cite{AtteneF06}) varying from 10 to 100.}
\end{figure}

As a mesh model can be imported from different sources, it is possible that there can be noise in the data. To test the algorithm for noisy inputs,  ReMESH 2.1 \cite{AtteneF06}, a mesh editing software, is used to generate them. Noise is varied by distributing the Gaussian noise over the model in the normal direction and it is performed by increasing the percentage of bounding ball radius (BBR) of the mesh model \cite{AtteneF06}. Figure \ref{fig:noisy inputs} shows the results for models with varying noise specified by BBR. For BBR =  10 to 100, Figures \ref{fig:noisy inputs}(a)-(d)) show that the algorithm has recognised almost all the features correctly.

We can note that the predicted labels do not change significantly for low noise values (up to BBR of 100). However, in high value of noise (BR $>$ 100), some of the predicted classes are different from the original output. This is due to the fact that low value of noise does not affect the Gauss map significantly. For higher values of noise, the orientation of the facets gets distorted. This, in turn, affects the Gauss map and hence the signature to be significantly different from the original one thereby causing a misrecognition. To the best of our knowledge, no existing methods for feature recognition can handle data with noise. 

\subsection{Comparison with other works}
The FeatureNet \cite{ZHANG201812} is a publicly available dataset which contains a selected set of 24 commonly occurring features. FeatureNet uses 3D CNN which is a deep neural network and requires 390 minutes of training time and has a huge number of hyperparameters. The voxel resolution is 64x64x64. The recognition accuracy achieved was 97.4\%. Our machine learning approach gives a much faster recognition (total time of 3.19 seconds) with better accuracy.

FeatureNet is also modeled based on single feature learning and it requires feature extraction as in our method to perform multi-feature recognition. In practical cases, most of the models contain multiple features in it. Hence it is required to first extract the features before recognising it. The algorithm proposed by us has heavily reduced the memory requirement since the features have been reduced to a 1-dimensional vector [102 x 1]. Gauss map based approach also resulted in a much faster run time while training as well as testing and also showed improved performance in terms of accuracy. Time for feature extraction is not included in the running time similar to excluding the time for segmentation in  \cite{ZHANG201812}. Table \ref{table:FeatureNet_compare} summarizes the comparison of our approach with FeatureNet \cite{ZHANG201812}.

It may be noted that the run time for Msvnet \cite{Shi2020} for 64X64X64 configuration is 871.23 min for whole dataset. Moreover, they also employ 'number of sectional views' as parameter which they arrive at using trial and error. This approach also needs to perform segmentation of the model and then recognise the features. It should be mentioned that their near-optimal performance accuracy (98.33\%) with a split of 80-20 is little high than ours (97.9\%) with a split of 70-15-15. The STL files were also converted to voxels whereas we don't need conversion.

In either of the methods, there is no demonstration of their performances with varying noise levels in the input data. Also, deep networks are usually run on a GPU with high memory (for e.g., \cite{Shi2020} uses 128GB RAM with 2080Ti GPU) whereas ours can be run on CPU itself.

\section{Conclusion} \label{sec_conc}

In this paper, we proposed random forest as a classifier for the recognition of machining features. The  discrete Gauss map approach for shape signature helped in not only reducing the memory requirement but also resulted in very good accuracy to that of modern 3D CNN approaches in much lesser training time. This approach also paved the way to handle complex/interacting features. It was demonstrated that the discrete Guass map enable to handle certain level of noise in the input data. Overall, it was shown that the problem of feature recognition could be addressed using simpler machine learning approach than using deep networks, and achieve similar accuracy with minor trade-off. Our method also suffers from the same drawback as the others \cite{ZHANG201812, Shi2020} in that a multi-feature model needs segmentation to single features. In future, reducing the mis-classification and improving the accuracy further would be the focus. 
 
\bibliographystyle{cag-num-names}
\bibliography{FeaRecog_arxiv}

\end{document}